\documentclass[times,review,preprint,authoryear]{elsarticle}

\usepackage{medima}
\usepackage{framed,multirow}
\usepackage{latexsym}

\usepackage{amsmath,amssymb,amsfonts}
\usepackage{latexsym}

\usepackage{algorithmic}
\usepackage{graphicx}
\usepackage{textcomp}
\usepackage{mathrsfs}
\usepackage{makecell}
\usepackage{makeidx}  
\usepackage{url}
\usepackage{longtable}
\usepackage{multirow}
\usepackage{booktabs}
\usepackage{url}
\usepackage{xcolor}
\usepackage[citecolor=green]{hyperref}

\usepackage{mathrsfs}
\usepackage{makecell}
\usepackage[linesnumbered,ruled,vlined]{algorithm2e}
\usepackage{diagbox}

\definecolor{red}{rgb}{.8,.349,.1}

\journal{Medical Image Analysis}

\begin{document}
	
	\verso{Xin Yang and Yuhao Huang \textit{et~al.}}
	
	\begin{frontmatter}
		
		\title{Searching Collaborative Agents for Multi-plane Localization in 3D Ultrasound}%
		
		\author[1,2]{Xin \snm{Yang}\fnref{fn1}}
		\author[1,2]{Yuhao \snm{Huang}\fnref{fn1}}
		\fntext[fn1]{The two authors contribute equally to this work.}
		\author[1,2]{Ruobing \snm{Huang}}
		\author[1,2]{Haoran \snm{Dou}}
		\author[1,2]{Rui \snm{Li}}
		\author[1,2]{Jikuan \snm{Qian}}
		\author[1,2]{Xiaoqiong \snm{Huang}}
		\author[1,2]{Wenlong \snm{Shi}}
		\author[1,2]{Chaoyu \snm{Chen}}
		\author[3]{Yuanji \snm{Zhang}}
		\author[3]{Haixia \snm{Wang}}
		\author[3]{Yi \snm{Xiong}\corref{cor1}}
		\ead{13352995536@163.com}
		\author[1,2]{Dong \snm{Ni}\corref{cor1}}
		\ead{nidong@szu.edu.cn}
		\cortext[cor1]{Corresponding author.}
		
		
		\address[1]{National-Regional Key Technology Engineering Laboratory for Medical Ultrasound, School of Biomedical Engineering, Health Science Center, Shenzhen University, Shenzhen, China}
		\address[2]{Medical Ultrasound Image Computing (MUSIC) Laboratory, Shenzhen University, Shenzhen, China}
		\address[3]{Department of Ultrasound, Luohu People’s Hospital, Shenzhen, China}
		
		\received{*****}
		\finalform{*****}
		\accepted{*****}
		\availableonline{*****}
		\communicated{*****}

		\begin{abstract}
			3D ultrasound (US) has become prevalent due to its rich spatial and diagnostic information not contained in 2D US. Moreover, 3D US can contain multiple standard planes (SPs) in one shot. Thus, automatically localizing SPs in 3D US has the potential to improve user-independence and scanning-efficiency. However, manual SP localization in 3D US is challenging because of the low image quality, huge search space and large anatomical variability. In this work, we propose a novel multi-agent reinforcement learning (MARL) framework to simultaneously localize multiple SPs in 3D US. Our contribution is four-fold. First, our proposed method is general and it can accurately localize multiple SPs in different challenging US datasets. Second, we equip the MARL system with a recurrent neural network (RNN) based collaborative module, which can strengthen the communication among agents and learn the spatial relationship among planes effectively. Third, we explore to adopt the neural architecture search (NAS) to automatically design the network architecture of both the agents and the collaborative module. Last, we believe we are the first to realize automatic SP localization in pelvic US volumes, and note that our approach can handle both normal and abnormal uterus cases. Extensively validated on two challenging datasets of the uterus and fetal brain, our proposed method achieves the average localization accuracy of 7.03$^{\circ}$/1.59\emph{mm} and 9.75$^{\circ}$/1.19\emph{mm}. Experimental results show that our light-weight MARL model has higher accuracy than state-of-the-art methods.
		\end{abstract}
		
		\begin{keyword}
			\MSC *****\sep *****\sep *****\sep *****
			\KWD \\Neural Architecture Search\\Reinforcement Learning\\Collaborative Agents\\3D Ultrasound\\Plane Localization
		\end{keyword}
		
	\end{frontmatter}
	\section{Introduction}
	\label{introduction}
	
	\begin{figure*}[t]
		\centering
		\includegraphics[width=1.0\linewidth]{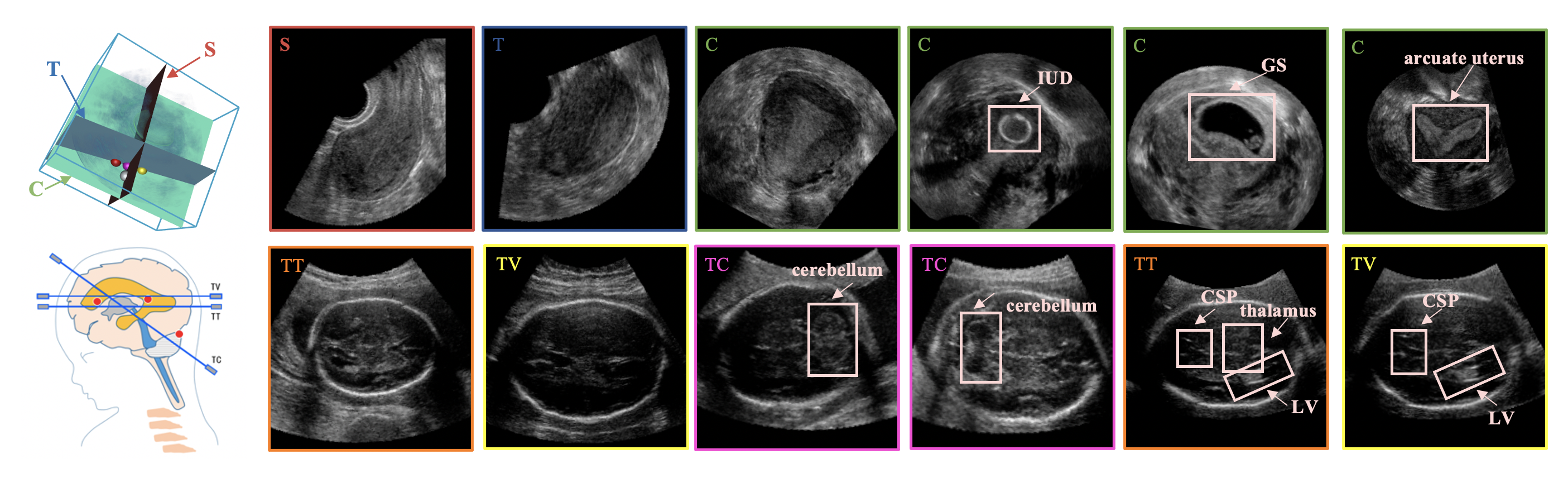}
		\caption{Visualization of the targeted SPs in 3D US \emph{(left to right)}. \emph{Top row (uterus):} Spatial layout of uterine SPs (the red, yellow, purple and white dots are two endometrial uterine horns, endometrial uterine bottom and uterine wall bottom). One typical example of the mid-sagittal (S), the transverse (T) and the coronal (C) plane in the uterus. To illustrate the existence of special or abnormal cases in this dataset, we also show examples with IUD, GS or arcuate uterus. Note that the appearance of the coronal plane can change dramatically in these cases. \emph{Bottom row (fetal brain):} Schematic diagram of anatomy (three red dots are the genu of corpus callosum, splenium of corpus callosum and cerebellar vermis from left to right). One typical example of the transthalamic (TT), transventricular (TV) and the transcerebellar (TC) plane in the fetal brain. We show the TC plane of another subject. It proves that the fetal brain can have $180^{\circ}$ orientation difference (note the cerebellum can appear on both sides of the image.) Also, note that the TT and TV planes are spatially close, their appearance can be extremely similar (e.g. the two right-most sub-figures) and add more difficulty to the task.}
		\label{fig:intro}
		\vspace{-0.4cm}
	\end{figure*}

	Manipulating ultrasound (US) probes to accurately localize standard planes (SPs) is vital for subsequent biometric measurement and diagnosis during routine US examinations~\citep{bornstein2010basic,loughna2009fetal,moellers2018fetal}. SPs containing key anatomical structures are often manually selected by sonographers from 2D image sequences. However, this procedure is highly operator-dependent and time-consuming, since the SP position is very fine and a tiny probe movement may result in a large deviation~\citep{ni2014standard}. In comparison, the advance of 3D US technology enables multiple SPs to be contained in one shot, which has the potential to reduce operator-dependence and improve scanning efficiency. 3D US can also provide rich spatial and diagnostic information not contained in 2D US. Note that the coronal SPs of the uterus can only be reconstructed from 3D US and are important for assessing congenital uterine anomalies and Intra-Uterine Device (IUD)~\citep{wong2015three}. However, manually localizing SPs in 3D US is challenging because of the huge search space and large anatomical diversity. Therefore, an automatic approach to localize SPs in 3D US is desired to simplify the clinical process and reduce observer dependency.

	As shown in Fig.~\ref{fig:intro}, there remain several challenges for localizing SPs in the 3D US automatically. First, the same SPs often have high intra-class variation. For the uterus, the variation is partly caused by the presence of gestational sac (GS) and IUD in normal subjects. Furthermore, the presence of congenital uterine anomalies (e.g. rudimentary or bicornuate uterus) and other diseases (e.g. endometrial disorders) aggravate the problem largely. In the fetal brain, the high intra-class variation of SPs comes from the scale, shape and contrast difference caused by fetal brain development, and also the varying fetal postures. Second, the inter-class variation among different types of planes is often low. The differences between SPs and non-SPs, and even different SPs are often negligible, since they can contain the same or different anatomical structures with a similar appearance. For example, in fetal brain US, both cavum septum pellucidum (CSP) and lateral ventricle (LV) appear on TT, TV and other non-SPs, which may cause a high confusion for the algorithms. Third, SPs have different appearance patterns and fine definitions, which makes designing a general machine learning algorithm difficult. The fourth challenge lies in the varied spatial relationship among SPs. For example, three uterine SPs are usually perpendicular to each other. However, this is not always true due to the existence of anomalies and disorders.

	In this study, we propose a novel Multi-Agent Reinforcement Learning (MARL) framework to localize multiple SPs in 3D US simultaneously. The main contributions of our work are:

	\begin{enumerate}
		\item A robust and general approach to detect multiple planes simultaneously in challenging 3D US datasets. Validation experiments showed that it can pinpoint SPs in different datasets containing organs with large anatomical variations.
		\item An MARL framework with a novel recurrent neural network (RNN) based collaborative module. This design encourages collaboration in learning to identify multiple targets, whose relative spatial locations are essential for each accurate predictions.
		\item Neural architecture search (NAS) to automatically design the optimal models for RL agents, and obtain the auto-searched RNN module, respectively. The searched models are light-weight while yielding higher accuracy.
		\item To the best of our knowledge, this is the first work that addresses automatic SPs localization in pelvic US volumes, moreover, the proposed model is capable of handling both normal and abnormal cases.
	\end{enumerate}

	Therefore, identifying SPs in these two scenarios are both complicated and challenging. Validation experiments demonstrated that the proposed framework can achieve high accuracy in both tasks. Next, we discuss related works in Section~\ref{Related_works}. The proposed framework is explained in Section~\ref{METHODOLOGY}. Section~\ref{experiment result} demonstrates experiments and the results, and Section~\ref{conclusion} concludes our work.
	
	\section{Related Works}
	\label{Related_works}
	\subsection{Standard Planes Detection in 2D US Images}
	In one of the earliest studies,~\cite{zhang2012intelligent} proposed an automatic detection system for early gestational sac SP using two cascaded AdaBoost classifiers in a coarse-to-fine manner. The other methods~\citep{ni2013selective, ni2014standard, yang2014standard} classify the SP in the US videos by detecting the key anatomical structures in the fetal abdomen using different classical machine learning algorithms, e.g. random forests, AdaBoost. 
	Recently, convolutional neural networks (CNNs) based classification methods were employed for 2D SP detection.
	~\cite{chen2015standard} first constructed the classifier based on transfer learning to detect the fetal SPs.
	~\cite{baumgartner2016real, Baumgartner2017SonoNetRD} detected SPs based on a classification model and further located the anatomical structures of SP using unsupervised and weakly-supervised learning, respectively.
	~Gated attention mechanism~\citep{schlemper2018attention} technology was then proposed and incorporated into the classification network for improving the performance of SP detection.
	Additionally, some efforts utilized RNN in the detection system to capture the temporal-spatial information in 2D US videos~\citep{chen2017ultrasound,lin2019multi}.

	Although the aforementioned methods are effective in detecting SPs in 2D US, their tasks are essentially different from SPs localization in 3D US. 
	In 2D US, plane localization is considered as a frame search task in video sequences with countable frames and thus the classification-based methods can handle this well.
	However, compared with 2D US, the 3D US contains countless slices, which makes the classification-based approaches intractable and brings exponentially growing false positives.
	Besides, localizing planes in 3D US requires pinpointing the exact locations and navigating to the correct orientation in the huge space simultaneously, which is more complex than just locating the frame index in 2D US.
	
	\subsection{Standard Planes Localization in 3D Volumes}
	Automatic solutions for plane localization in 3D volumes can be roughly classified into two types: 1) Supervised Learning (SL) based methods such as registration, regression and classification, and 2) Reinforcement Learning (RL) based approaches. 
	
	\subsubsection{SL Methods}~\cite{lu2011automatic} proposed a boosting-based classification approach for automatic view planning in 3D cardiac MRI volumes.~\cite{chykeyuk2013class} developed a random forest-based regression method to extract cardiac planes from 3D echocardiography.~\cite{ryou2016automated} proposed a cascaded algorithm to localize SPs of fetal head and abdomen in 3D US. An SL based landmark-aware registration solution was also employed to detect fetal abdominal planes~\citep{lorenz2018automated}. Two regression methods were applied to localize the TT and TV planes in the fetal brain~\citep{li2018standard}, and the abdominal circumference (AC) planes~\citep{schmidt2019offset}. 
	Although effective, most of these methods are specially designed for one single organ or can only localize one plane at once. 
	Furthermore, they try to directly learn the mapping from high-dimension volumetric data to the low-dimension abstract features (i.e., plane parameters), which is complex and difficult to train.  
	Besides, the spatial dependency among different targets, which is implicitly contained in the volume, can not be well distilled by previous methods due to their \textit{grid-wise} 3D convolution kernels, rather than the \textit{plane-wise} 2D ones in our method.
	

	

	\subsubsection{RL Methods} 
	\label{Sec:RL}
	In an RL framework, an agent interacts with the environment to learn the optimal policy that can maximize the accumulated reward. It is similar to the behavior of sonographers as they manipulate ($action$) the probe ($agent$) to scan the organ ($environment$) while visualizing the intermediate planes on the screen ($state$) till the SP is acquired ($reward$ and $terminal$ $state$).~\cite{alansary2018automatic} were the first to utilize RL in SP localization for 3D MRI volumes.~\cite{dou2019agent} proposed to equip the RL framework with a landmark-aware alignment module for warm start and an RNN based termination module for stopping the agent adaptively during inference. These approaches achieved high accuracy and showed great potential in addressing the issues. However, they are both limited in the sense that they can only locate one SP each time. It can lead to longer running time and sub-optimal performance when multiple SPs are desired and blind in relationships between targets. Furthermore, lacking target-specific model design, these methods might be unable to capture the complex appearance patterns of different SPs using the same exact model.
	
	MARL system has been proposed and applied in a variety of domains~\citep{bucsoniu2010multi}. Recently,~\cite{vlontzos2019multiple} proposed a collaborative MARL framework for detecting multiple landmarks in MRI volumes. In their framework, the agents can only communicate with each other by sharing the low-level representations in a transfer learning manner, which can not explicitly represent the spatial relationships among SPs.
	Our work, for the first time, exploits this strong dependency explicitly through an RNN collaborative module by sharing the high-level decision information of agents. Details are introduced in Section~\ref{METHODOLOGY}.

	
	\subsection{Neural Architecture Search}
	
	Designing suitable neural network architectures for specific tasks requires substantial knowledge and expertise. Moreover, the optimal heuristics in designing this remains an open question. Therefore, Neural Architecture Search (NAS) that endeavors to address this issue, has been utilized in a wide variety of applications for performance improvement or obtaining lightweight models~\citep{elsken2018neural,wistuba2019survey,ren2020comprehensive}. Because of discrete optimization, early search methods (e.g. RL-based NAS~\citep{zoph2016neural,baker2016designing,zoph2018learning}, evolutionary-based NAS~\citep{real2017large, xie2017genetic,real2019aging,real2019regularized}) evaluate many searching architectures from scratch and often demand large computation resources. On the contrary, gradient-based approaches are more efficient because they adopt a hypernetwork instead of evaluating countless architectures.~\cite{liu2018darts} first proposed the differentiable architecture search (DARTS) by establishing a continuous architecture search space.~\cite{dong2019searching} introduced the gradient-based differentiable architecture sampler (GDAS) method which only updates the sub-graph sampled from the supernet in each iteration to accelerate and stabilize the learning process. To automatically and adaptively discover a customized model for our task, we propose to utilize NAS, in specific, GDAS to search the architectures for each of the RL agents, which enables each agent to adapt to its own target freely without requiring prior knowledge. We also use GDAS to find an efficient RNN model, with a similar amount of parameters but achieving better performance.

	To address the multi-plane localization tasks in 3D US, our previous MICCAI version method~\citep{huang2020searching} proposed a MARL based framework to detect three SPs in 3D uterine US volumes simultaneously. Moreover, we equipped the framework with an RNN based collaborative module to catch the latent spatial relationships among multiple planes. Specifically, for the RL agents, we adopted NAS methods to search automatically, while for the collaborate module, we simply used a manually-design RNN. In this work, we extend our previous method and further propose to utilize a searched RNN in the collaborative module. Besides, instead of validating our method on uterus dataset only, we add a large 3D fetal brain US dataset for validation. Experiments show that the searched models outperform our previous methods and only require a similar amount of parameters.

	We integrate the strength of both MARL and NAS to build a general framework for SPs localization in 3D US. Meanwhile, the RNN based collaborative module can actively investigate the spatial dependency among targets, which is a general approach for collaborating the RL agents. Our proposed methodology is introduced in detail in the next section.

	\section{Methodology}
	\label{METHODOLOGY}
	
	\begin{figure*}[t]
		\centering
		\includegraphics[width=1.0\linewidth]{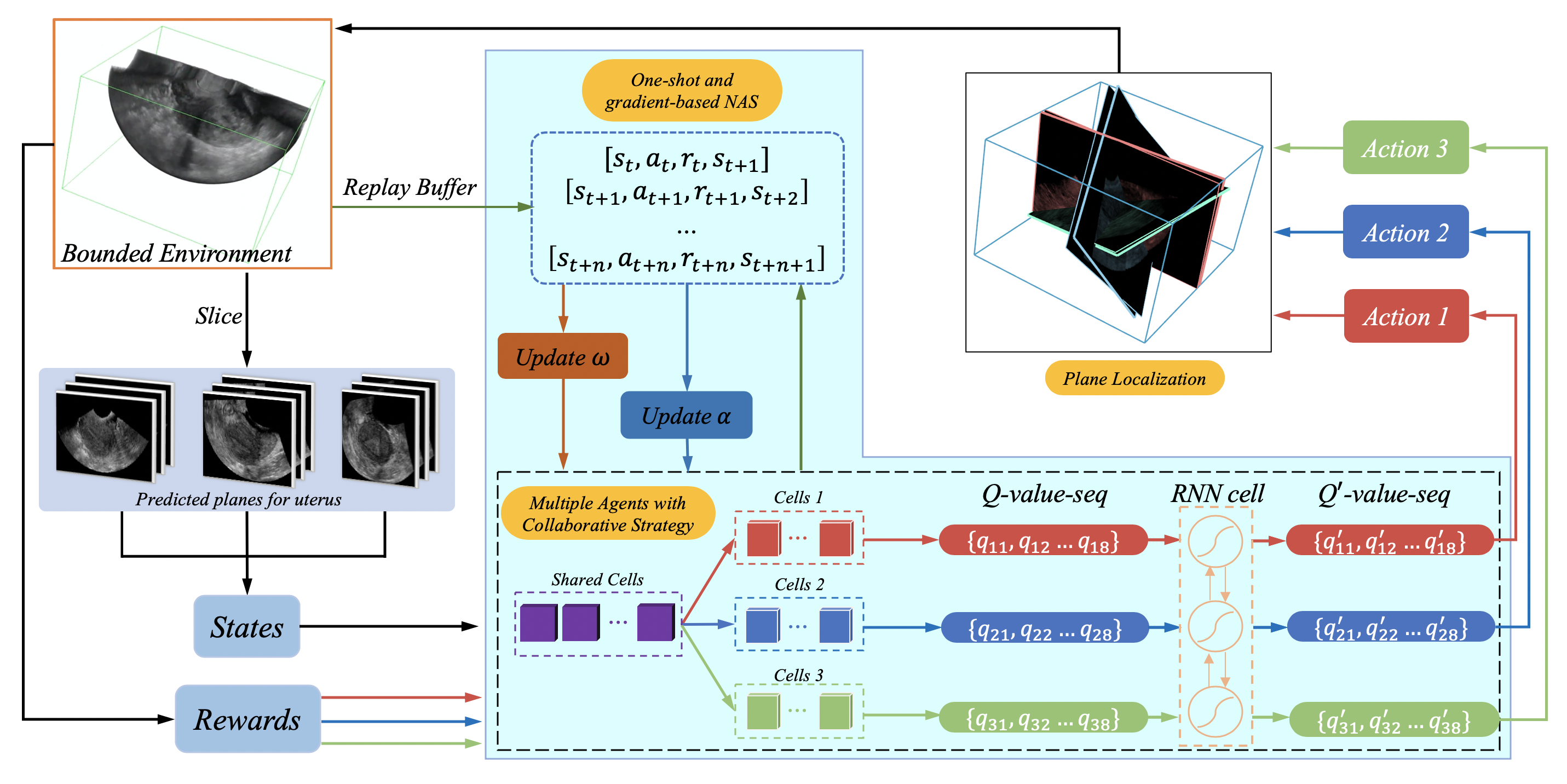}
		\caption{Overview of the proposed framework. Here, we use the uterine volume as an example to illustrate the whole learning process.}
		\label{fig:framework}
		\vspace{-0.4cm}
	\end{figure*}
	
	Fig.~\ref{fig:framework} shows the workflow of our proposed method. Our proposed MARL framework can localize multiple SPs in 3D US at the same time. In this work, we aim at simultaneously localizing 1) mid-sagittal (S), transverse (T) and coronal (C) planes in volumes and 2) trans-thalamic (TT), trans-ventricular (TV)and trans-cerebellar (TC) planes in fetal brain US volumes. We first adopt a landmark-alignment module to provide effective initialization and strong spatial bounds for agents against the noisy 3D US~\citep{dou2019agent}. Four landmarks of uterus and three landmarks of fetal brain used for registration are shown in Fig.~\ref{fig:intro}. We further equip an RNN based collaborative module among agents. It can help agents share their spatial information and action policies, thus learning the spatial relationship among planes effectively. Then, we adopt one-shot and gradient-based GDAS~\citep{dong2019searching} to search 1) optimal agents for different SPs and 2) the RNN based collaborative module.

	\subsection{MARL Framework for Multi-plane Localization}
	
	To simultaneously localize multiple SPs in uterine and fetal brain US respectively and address the non-stationary of environment happens in MARL models, we establish communication between agents and propose a collaborative MARL framework. To better understand the MARL framework, we first introduce the Single Agent RL (SARL) simply. In a typical SARL system, the agent learns a policy (i.e. action sequence) during the interaction with the environment to maximize the accumulated reward of the task~\citep{kaelbling1996reinforcement}. While in a MARL system, multiple agents are created and the system endeavors to maximize the total reward obtained by all the agents~\citep{foerster2016learning,gupta2017cooperative}. 
	
	The MARL framework can be defined by the~\emph{Environment},~\emph{States},~\emph{Actions},~\emph{Reward} and~\emph{Terminal States}. To formalize the considered SP detection task as a MARL system, we define the~\emph{Environment} as the 3D US volume (see Fig.~\ref{fig:framework}). The three targeted SPs ($P_1,P_2,P_3$) are searched using three agents: $agent_{k,\; k=1,2,3}$.
	The~\emph{States} are defined as the last nine planes predicted by three agents, with each agent obtaining three planes. This setting can provide rich state information for the agents while keep the learning speed~\citep{mnih2015human,dou2019agent}.
	With the plane normal vector $(\cos(\zeta),\cos(\beta),\cos(\phi))$ and its distance from the volume center $d$, the plane function in Cartesian coordinate system can be formulated as: $\cos(\zeta)x+\cos(\beta)y+\cos(\phi)z+d=0$. We then define the~\emph{Actions} that the agents can take as \{$\pm$\emph{a}$_{\zeta}$, $\pm$\emph{a}$_{\beta}$, $\pm$\emph{a}$_{\phi}$, $\pm$\emph{a}$_{d}$\}. The plane parameters of the $P_1,P_2,P_3$ can be updated accordingly to obtain new planes (e.g. $\zeta \pm \alpha_\zeta$). The ~\emph{Reward}, $R_{k} \in \{-1, 0, +1\}$, for each $agent_{k}$ can be calculated by:
	\noindent
	\begin{equation}
		\emph{R}_{k}=\emph{sgn}(\emph{D}{_k^{t-1}}-\emph{D}{_k^t}),
		\label{equ0}
	\end{equation}
	where $sgn$ is a sign function and $t$ represents the iteration step in agent interacting. $\emph{D}{_k^t}$ represents the difference between the plane parameters of plane~\emph{P}${_k^t}$ and that of its corresponding ground truth~\emph{P}${_k^g}$. Specifically, the difference is defined as the $Euclidean$ $distance$: $\emph{D}{_k^t} = \|$\emph{P}${_k^t}$$-$\emph{P}${_k^g}$$\|_2$. For the~\emph{Terminal States}, we choose a fixed-steps strategy to stop the agent-environment interaction and treat the last step prediction as the final output.

	One could imagine that there exist numerous combinations of states and actions before the agents reach the final states. Therefore, instead of storing all the state-action values (Q-values) in a table in classical Q-learning~\citep{watkins1992q}, we opt for a strategy like deep Q-network (DQN)~\citep{mnih2015human} where a CNN is used to model the relationship among states, actions and Q-values. In specific, we adopt the double DQN (DDQN)~\citep{van2016deep} method, for mitigating the upward bias caused by DQN and better stabilizing the training process. 
	Besides, we used a prioritized replay buffer~\citep{schaul2015prioritized} to store the data sequences containing states $s$, actions $a$, rewards $r$ and the states of the next step $s'$ in the agent-environment interaction, which can remove data correlations and improve sampling efficiency. Specifically, the $i$-$th$ sequence element with high error will be sampled from the buffer preferentially, and its sampling probability can be calculated as $P_{i}=\frac{e_i^{p}+\delta}{\sum_{k}(e_k^{p}+\delta)}$, where $p=0.6$ controls how much prioritization is used and $\delta=0.05$ is set to adjust the error $e_{i}$.
	Besides, we also adopted importance-sampling weights for correcting the bias caused by the change of data distribution in prioritized replay similar to~\citep{schaul2015prioritized}.
	Our MARL system is trained using the following loss function: 
	\noindent
	\begin{equation}
		\mathcal{L} = E[(r+\gamma Q^*(s',\mathop {argmax} \limits_{a'}Q(s',a';\omega);\tilde{\omega})-Q(s,a;\omega))^2],
		\label{equ1}
	\end{equation}
	where $\gamma\in[0,1]$ is a discount factor to weight future rewards. The $Q$ network $Q(\omega)$ is used to obtain the Q-values of the current step and the actions $a'$ in the next step. $Q^*(\tilde{\omega})$ represents the target $Q$ network leveraged to estimate the Q-values of the next step. $\omega$ and $\tilde{\omega}$ are the network parameters of $Q$ network and target $Q$ network. $\epsilon - greedy$ algorithms are utilized for the action selection to balance the exploration and exploitation during training~\citep{mnih2015human}. Normally, the Q-values will be output by the fully connected layers and guide the actions of agents to update the plane parameters. To learn the latent spatial relationship among planes, we equip an RNN based agent collaborative module after the fully connected layers. Thus, the agents will take actions following the calibrated Q-values output by the RNN module instead of the vanilla Q-values.
	
	\begin{figure*}[t]
		\centering
		\includegraphics[width=1.0\linewidth]{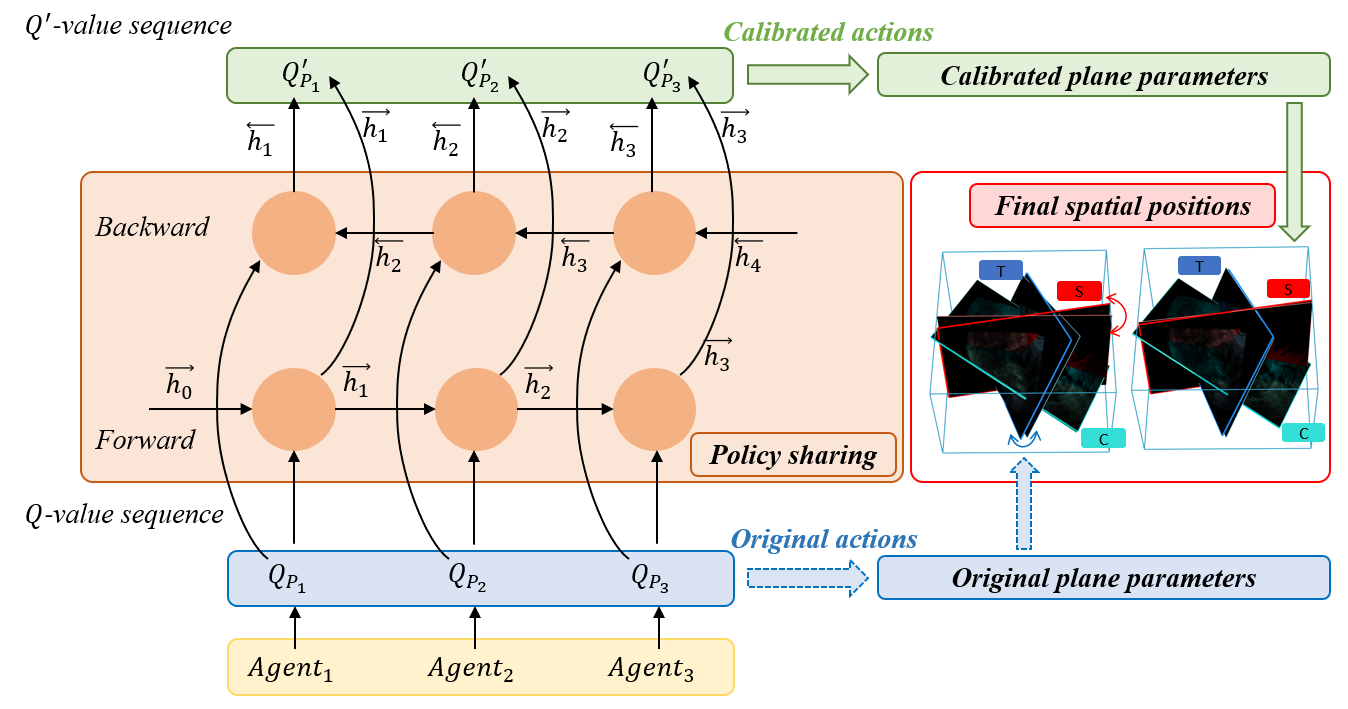}
		\caption{The proposed RNN collaborative module. The blue block and dotted arrows indicate the vanilla plane-updating process which is based on the independently predicted Q-value sequence. The orange block is the RNN cell that makes the agents share policies. It takes the original Q-value sequence as inputs and outputs the calibrated Q'-values (green block). The proposed method uses the updated Q'-values to choose actions and update plane parameters (solid green arrow) to decide the final spatial position of planes (red block). In the red block, we show the spatial relationships among planes in uterus US volume obtained by models without and with RNN-based agent collaborative module ($Left:$ G-MARL and $Right:$ Ours). The red and blue arrows in $left$ figure indicate the spatial differences between ground truth and prediction pairs ($Red$: S plane and $Blue$: T plane).}\label{fig:rnn_module}
		\vspace{-0.45cm}
	\end{figure*}
	
	\subsection{RNN based Agent Collaborative Module}
	Fig.~\ref{fig:intro} shows the most typical spatial relationships among planes in the uterus and the fetal brain, and these relationships vary within certain ranges.
	Thus, learning such spatial information could boost the robustness and the accuracy of an automatic multi-plane localization model. However, the previous methods have not considered such latent spatial information. Broadly, there can be divided into two categories and each with its limitations:
	
	
	\begin{enumerate}
		\item[i)] The previous RL-based methods for plane localization~\citep{alansary2018automatic,dou2019agent} used the Single-Agent RL (SARL) framework, in which each agent was trained separately and thus the information within and among planes cannot be fully utilized. 
		\item[ii)] The collaborative MARL framework~\citep{vlontzos2019multiple} for detecting multiple landmarks. It can learn a common knowledge of planes through parameter sharing in the convolutional layers. However, this is not an intuitive strategy for learning spatial information among planes.
	\end{enumerate}

	In this work, we model such spatial relations explicitly using a novel RNN based agent collaborative module to improve the accuracy of plane localization. In specific, instead of directly outputting the Q-values through fully connected layers without communication like the established MARL frameworks~\citep{bucsoniu2010multi} (solid blue arrow in Fig.~\ref{fig:rnn_module}), an RNN cell with bi-directional structure is proposed to be inserted among the agents (see Fig.~\ref{fig:framework} and Fig.~\ref{fig:rnn_module}). The main reason for applying a bi-directional structure is that it can combine forward and backward information, and thus realizes information flowing and policy sharing among multiple agents. Therefore, it can enhance communication among agents.
	
	Agents can collaborate to benefit each other in the learning of \textit{plane-wise} spatial features. The Q-values in recurrent form are the collaborate media, which encode the $states$ and $actions$ (i.e., $Q=\hat Q(s, a)$, where $\hat Q(\cdot)$ represents the $Q$ network), and explicitly reflect the high-level knowledge including plane spatial positions and trajectory policy information.
	It is noted that such high-level decision information will determine the final spatial position of planes (see Figure~\ref{fig:rnn_module}), thus they explicitly contain the spatial relationship among planes.
	Therefore, the agents can learn the plane relationships explicitly through high-level knowledge sharing, and take the calibrated actions based on Q'-values.
	Specifically, at step $t$, the Q-value sequence set is denoted as $Q_{t}=(Q_{P_{1}},Q_{P_{2}},Q_{P_{3}})^\mathrm{T}$, where the Q-value sequence $Q_{P_{i}}$ for plane $P_{i}$, which contains 8 Q-values (according to the \emph{Actions} space), can be written as: $Q_{P_{i}}=\{q_{i1}, q_{i2}, q_{i3}..., q_{i8}\},~i=1, 2, 3$. 
	The Q-value sequence $Q_{P_{i}}$ of each plane is then passed to an RNN with bi-directional structure as its hidden-state of each time-step. Then the RNN module outputs a calibrated Q-value sequence set $Q'_{t}=(Q'_{P_{1}},Q'_{P_{2}},Q'_{P_{3}})^\mathrm{T}$ and thus exploits full knowledge of all the detected planes, formally, we defined:
	\noindent
	\begin{equation}
		Q'_{t} = \mathcal{H}(Q_{t}
		,\stackrel{\rightarrow}{h^{\tilde{t}-1}},\stackrel{\leftarrow}{h^{\tilde{t}+1}};\theta),
		\label{equ3}
	\end{equation}
	where $\stackrel{\rightarrow}{h^{\tilde{t}}}$ and $\stackrel{\leftarrow}{h^{\tilde{t}}}$ are the forward and backward hidden sequence, respectively. $\mathcal{H}$ is the hidden layer function including linear and different activation operations, and $\theta$ represents the model parameters of the RNN. Then the agents take calibrated actions based on $Q'_{t}$ to update the plane parameters and decide the final positions of planes (solid green arrow in Fig.~\ref{fig:rnn_module}). In other words, the actions taken by each agent are based on not only its own action policy and spatial information, but also the policies and information shared by other agents. As also can be seen in the red block of Figure~\ref{fig:rnn_module}, without the RNN module, though the C plane obtains a satisfactory result, the predicted S and T planes are far away from their targets ($left$ part). While after capturing spatial relationships using the RNN module, the spatial relationships among planes can be well constrained and optimized, thus making all predicted planes very close to their targets ($right$ part).
	
	\subsection{GDAS based Search Module}
	
	\begin{figure}[t]
		\centering
		
		\includegraphics[width=1.0\linewidth]{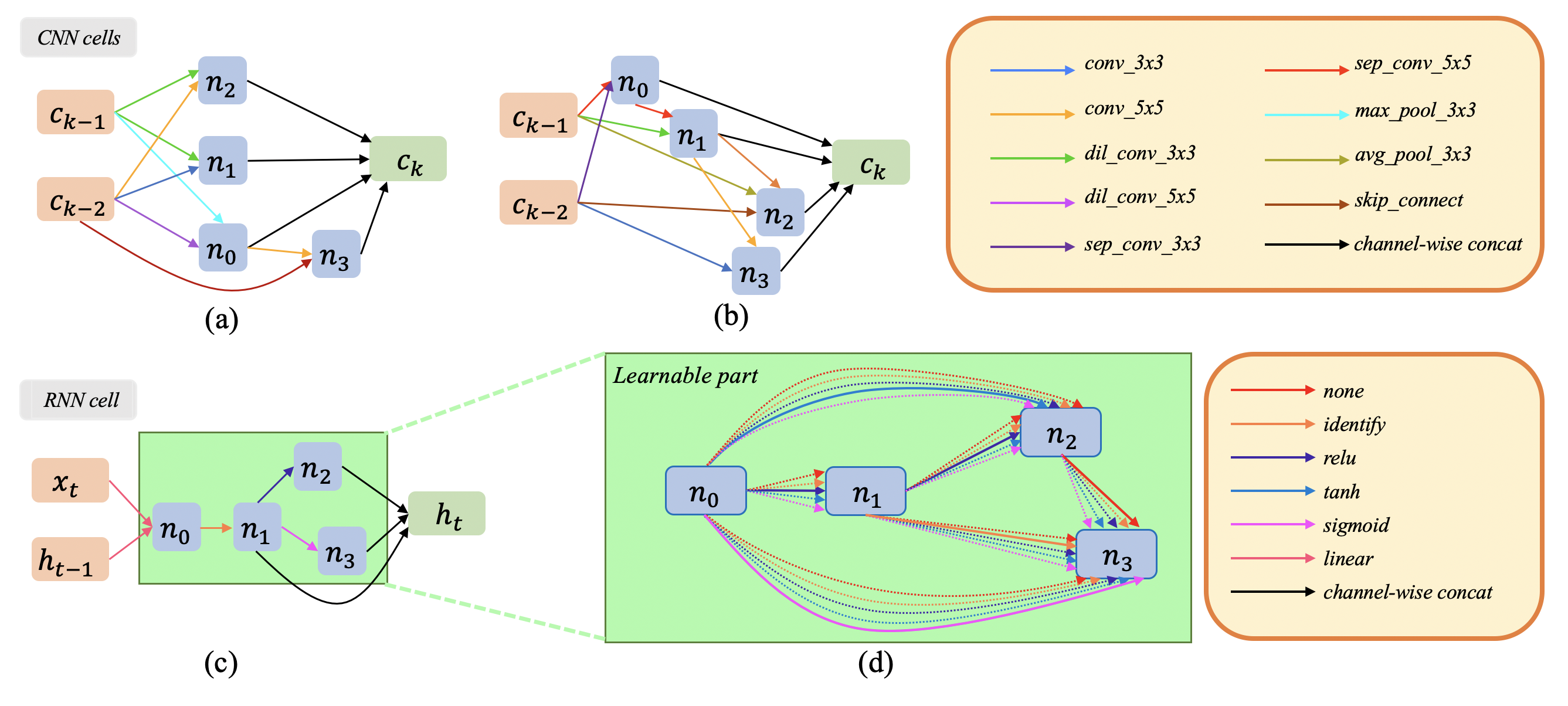}
		\caption{Details of CNN and RNN cells. (a) and (b) are two typical examples of CNN cells. (c) and (d) show one typical example of RNN cell and its learnable part, respectively. Different colored arrows represent different operations defined in the set of candidate operations. In the learnable part, the solid arrows form the sub-graph sampled from the supernet (represented by dotted arrows)}\label{fig:cnn_rnn}
		\vspace{-0.45cm}
	\end{figure}
	
	
	In deep RL, designing a suitable neural network architecture of the agent is crucial in achieving good learning performance~\citep{li2017deep}. Instead of using the same CNNs (like VGG) to localize different planes~\citep{alansary2018automatic,dou2019agent}, designing plane-specific, task-adapting and inter-tasks balanced neural network architectures for different planes enable more flexibility and might benefit the learning. However, this process is time-consuming, inconvenient to extend, and highly dependent on the expert experience. Therefore, we explore to take advantage of NAS methods to design plane-specific networks automatically. To the best of our knowledge, it is the first time that NAS method is adopted to search the network architecture of agent in RL (i.e. CNN). Similarly, the architectures of the RNN in the agent collaborative module may also influence its ability to extract the shared spatial information and action policies. Hence, we utilize similar technique in designing a suitable RNN model.

	Both RL and NAS require long training time and are hard to train even training them separately. Jointly training them thus faces more serious challenges. To address this issue, we explore to adopt one-shot and gradient-based NAS, i.e. GDAS, for searching both the CNN and the RNN to save training time and achieve a satisfactory performance. Here, one-shot and gradient-based methods represent that instead of training countless candidate sub-networks from scratch, we only need to train one supernet which is updated by gradient descent. Then, we can sample lots of sub-graph from the supernet for subsequent performance estimation.

	Fig.~\ref{fig:framework} illustrates the whole designed architecture. The details of its components are shown in Fig.~\ref{fig:cnn_rnn}. The search process of both the agent (CNN) and the collaborative module (RNN) can be defined by the following elements: search space, search strategy and performance estimation strategy. Simply, search space represents the whole pre-defined network structure to be searched (in this paper, i.e. supernet) and the candidate operations. The search strategy defines how to search and update the network structure. Besides, a performance estimation strategy is defined as the method for evaluating the performance of searched architectures.
	
	\subsubsection{Search Space}
	
	Designing a large search space, including the number of layers, network branches, connection rules, candidate operations, hyper-parameters, etc., is time-consuming and may obtain an over-complex search space. In order to improve design efficiency and simplify the search space, we use the cell-based structure (like VGG, ResNet, etc.). It only needs to define several types of cells (including the number of nodes, filters and channels, the connection rules between nodes, etc.), which can be stacked to form the final search structure according to the connection rules of cells. The cell is a convolutional block and nodes in the cell represent the features maps. The basic design of the cells to construct the agent (i.e. CNN) and the RNN are shown in Fig.~\ref{fig:cnn_rnn}, where the arrows represent different operations.
	
	Similar to DARTS~\citep{liu2018darts}, the CNN has two kinds of convolutional cells including normal cell and reduction cell. In definition, the input and the output of normal cells have the same size, while reduction cells output pooled feature maps with doubled channels. Each CNN cell consists of 7 nodes, including two input nodes, four intermediate nodes and one output node. $c_{k-1}$ and $c_{k-2}$ are equal to the outputs of the forward two cells, and the output $c_{k}$ are defined as the channel-wise concatenation of nodes $n_{0}$ to $n_{3}$ (see Fig.~\ref{fig:cnn_rnn} (a)-(b)). In the proposed framework, as shown in Fig.~\ref{fig:framework}, three agents share 8 CNN cells (5 normal cells and 3 reduction cells) while each of them has 4 unique cells (3 normal cells and 1 reduction cell). Thereby, the agents can share the common low-level features while extracting unique high-level representations for their own tasks. The CNN cells consist of 10 candidate operations, including none, $3\times3$ and $5\times5$ convolutions, separable convolutions and dilated convolutions, $3\times3$ max pooling, $3\times3$ average pooling and skip-connection. In the validation stage, operations (except for $none$) with the top 2 weights are chosen from all previous nodes for each intermediate node in CNN cells (see Fig.~\ref{fig:cnn_rnn} (a)-(b)).

	The RNN module consists of only one cell with a bi-direction structure. The recurrent cell contains 5 candidate operations, including none, identity mapping and three activation functions (i.e. $tanh$, $relu$ and $sigmoid$). As shown in Fig.~\ref{fig:cnn_rnn} (d), our recurrent cell also consists of 7 nodes: two input nodes, four intermediate nodes and one output node. The first intermediate node $n_{0}$ is calculated based on two input nodes $x_{t}$ (input) and $h_{t-1}$ (the previous hidden state). The output current hidden state $h_{t}$ is defined as the average of the rest intermediate nodes (nodes $n_{1}$ to $n_{3}$). Different from the CNN cells, only one operation (except for $none$) with the highest weight is select for intermediate node $n_{1}$ to $n_{3}$ in the RNN cell (see Fig.~\ref{fig:cnn_rnn} (c)).

	\subsubsection{Search Strategy and Performance Estimation Strategy}
	
	For the search strategy, in DARTS, the sub-operations are optimized jointly, which means that it updates the whole supernet and all parameters in every iteration. Thus, DARTS might have two main disadvantages:
	\begin{enumerate}
		\item[i)] All connections (sub-operations) between nodes need to be calculated and updated in each iteration, resulting in heavy search time and memory consumption.
		\item[ii)] Optimizing all the sub-operations concurrently leads to competition among them, which may cause unstable training. For example, sometimes, different sub-operations may output opposite values with a summation that tends to zero. It will hinder the information flow between two connected nodes and thus make the learning process unstable.
	\end{enumerate}
	
	To accelerate and stabilize the learning process, GDAS uses the differentiable sampler to obtain the sub-graph from the supernet and only updates the sub-graph in each iteration. Specifically, GDAS uses the $argmax$ during the forward pass, and $softmax$ with $Gumbel$-$Max$ trick during the back propagation. The $Gumbel$-$Max$ trick used on $softmax$ can better approximate the realistic sampling operations. Specifically, when the training epoch increases and the softmax temperature parameter $\tau$ drops until $\tau \rightarrow 0$, the $Gumbel$-$softmax$ tends to $argmax$. We refer readers to~\citep{maddison2014sampling, dong2019searching} for more details. During search, the network weights $\omega$ and architecture parameters $\alpha$ are updated by gradient (Eq.~\ref{omega} and Eq.~\ref{alpha}) using prioritized weight based batch ($M_{p}$) and random-way based batch ($M_{r}$) data sampled from the replay buffer, respectively. 
	\begin{gather}
		\label{omega}
		\omega = \omega - \nabla_\omega\mathcal{L}_{s,a,r,s'\sim M_p}(q, q^*; \omega, \alpha),\\
		\alpha = \alpha - \nabla_\alpha\mathcal{L}_{s,a,r,s'\sim M_r}(q, q^*; \omega, \alpha),
		\label{alpha}
	\end{gather}
	where $\mathcal{L}$ is the MSE loss function. $q$ is the direct output of the $Q$ network $Q(\cdot)$, and $q^*$ is calculated based on both the $Q$ network and the target $Q$ network $Q^{*}(\cdot)$, as shown below:
	\begin{equation}
		q^{*} = r +\gamma Q^*(s', \mathop{argmax} \limits_{a'}Q(s', a'))
		\label{qq}
	\end{equation}
	
	For the performance estimation strategy, when NAS being applied in non-RL tasks, the selection of the searched model is usually decided on the loss function~\citep{elsken2018neural}. Specifically, after the loss becomes converged, one of the searched architectures will be chosen as the final designed model according to their performances on the validation set. However, the loss in RL is oscillated and difficult to converge. Therefore, we select the optimal architecture parameters $\alpha_{CNN}^*$ and $\alpha_{RNN}^*$ for both the CNN and RNN using the maximum of the accumulated rewards on the validation set in all the training epochs. Using these, we can then construct bespoke CNN for each agent and an ideal RNN for the collaborative module.

	\section{Experiments and Results}
	\label{experiment result}
	
	\subsection{Clinical Data Collection and Analysis}
	
	To validate the proposed framework, we used two challenging US datasets of different organs: the uterus and the fetal brain. The scans were obtained using a US system with an integrated probe. The local institutional review board approved the informed consent process.
	
	\emph{\textbf{Uterus Dataset.}} The uterus dataset consists of 683 volumes obtained from 476 patients. The average size of the volumes is 261$\times$175$\times$277 with isotropic voxel of size 0.5$\times$0.5$\times$0.5~$mm^3$. Four experienced radiologists annotated three SPs and four landmarks~(see Fig.~\ref{fig:intro}) in each volume under strict quality control. 
	We randomly split the dataset into 489, 50 and 144 volumes for training, validation and testing at the patient level.
	Note that this dataset contains nearly 26\% of abnormal cases, mainly including congenital developmental abnormalities, endometrial diseases, uterine wall diseases and cervical abnormalities. Meanwhile, the data that is considered `normal', also includes cases contain IUD (3\%) or pregnancy (23\%). This adds further obstacles in detecting the SPs.

	\emph{\textbf{Fetal Brain Dataset.}} The fetal brain dataset consists of 432 volumes acquired from different healthy subjects under trans-abdomen scanning protocol. The fetuses have a broad gestational age range: 19 to 31 weeks. As a result, the fetal brain structures have large variations in size, shape and appearance. The average volume size is 267$\times$205$\times$236 and with isotropic voxel size of 0.5$\times$0.5$\times$0.5~$mm^3$. Four sonographers with 5-year experience provided manual annotations of three landmarks~(see Fig.~\ref{fig:intro}) and three SPs for every volume. 
	We randomly split the dataset into 302, 30 and 100 volumes for training, validation and testing.
	

	\emph{\textbf{Evaluation Criteria.}} In this study, we used three criteria to evaluate the performance in terms of spatial and content similarities for plane localization. First, the dihedral angles between two planes (Ang) and the difference between their Euclidean distances towards the volume origin (Dis) are calculated to evaluate the spatial accuracy of the SP localization. Moreover, we used the Structural Similarity Index (SSIM)~\citep{wang2004image} to assess the content similarity between the planes.

	\begin{table*}[h]
		\caption{Spatial relationship and structural similarity among SPs}\label{data}
		\begin{center}
			\scriptsize
			\begin{tabular}{c|c|c|c|c|c|c}
				\toprule
				\multirow{4}*{Uterus}& Ang(S$\&$T) & Ang(S$\&$C) & Ang(T$\&$C) & Dis(S$\&$T) & Dis(S$\&$C)& Dis(T$\&$C)  \\
				\cline{2-7}
				&{84.14$\pm$7.12}&{84.15$\pm$10.64}&{82.02$\pm$10.33}&{4.09$\pm$3.51}&{4.19$\pm$3.33}&{3.42$\pm$2.61} \\
				\cline{2-7}
				& SSIM(S) &  SSIM(T) & SSIM(C)& SSIM(S$\&$T) &  SSIM(S$\&$C) & SSIM(T$\&$C) \\
				\cline{2-7}
				&{0.69$\pm$0.06}&{0.67$\pm$0.06}&{0.52$\pm$0.06}&{0.71$\pm$0.06}&{0.57$\pm$0.09}&{0.54$\pm$0.08}\\
				\hline
				\multirow{4}*{Fetal brain}& Ang(TT$\&$TV) & Ang(TT$\&$TC) & Ang(TV$\&$TC) & Dis(TT$\&$TV) & Dis(TT$\&$TC)& Dis(TV$\&$TC)  \\
				\cline{2-7}
				&{2.62$\pm$7.22}&{23.50$\pm$9.30}&{23.49$\pm$9.72}&{3.21$\pm$1.82}&{3.59$\pm$2.47}&{6.29$\pm$2.80} \\
				\cline{2-7}
				& SSIM(TT) & SSIM(TV) & SSIM(TC) & SSIM(TT$\&$TV) &  SSIM(TT$\&$TC) & SSIM(TV$\&$TC) \\
				\cline{2-7}
				&{0.63$\pm$0.09}&{0.63$\pm$0.08}&{0.61$\pm$0.09}&{0.85$\pm$0.07}&{0.76$\pm$0.08}&{0.75$\pm$0.07}\\
				\bottomrule
			\end{tabular}
		\end{center}
	\end{table*}
	
	\emph{\textbf{Data Distribution.}} The distribution of SPs and their appearance can be substantially different across subjects. This may result from the rapid changing of the brain structures due to maturation across the gestation, or is caused by the existence of pregnancy or uterine anomalies.
	To offer a peep into this quantitatively, we calculated the Ang($^{\circ}$), Dis(mm), and SSIM among pairs of SPs within the whole datasets used in this study (see Tab.~\ref{data}). Specifically, Ang(*\&*) and Dis(*\&*) reflect the average angle and distance between two planes in all same volumes. SSIM(*) evaluates the structural similarity of the selected SPs (e.g. the TT) across the whole dataset, which is calculated by taking the average of SSIM between SPs obtained from different volumes, and SSIM(*\&*) illustrates the average SSIM between two planes in all same volumes. In Tab.~\ref{data}, it can be observed that: (1) The spatial relationships among SPs are not totally determined. It can be seen that there exist large deviations among subjects; (2) The SSIM of the same type of SPs is relatively low (e.g. $SSIM(C) = 0.52\pm0.06$). This indicates that there exist large intra-class variations in the appearance of the target, which is another challenge of our task; (3) The SSIM between different types of SPs is comparable to that of the same type. This is especially true in TT and TV, which have a mean SSIM of 0.85. We conjecture that this may result from the natural definition of the TT and TV planes (see Fig.~\ref{fig:intro}), which are parallel and adjacent to each other.
	Note that the target organs can have large position variations (e.g. the fetal brain can be up-side-down due to its position in the uterus) in US volumes. All the volumes are pre-processed using landmark-based registration (details refer to~\cite{dou2019agent}). This process can roughly align the organ into a similar coordinate space.

	\subsection{Experimental Setup}
	
	The proposed framework is trained with Adam optimizer on an NVIDIA TITAN 2080 GPU. In the first stage of searching network architecture (about 3 days), we set the learning rates to 5e-5 for learning network weights $\omega$ and to 0.05 for learning the architecture parameters $\alpha$, respectively. In the second stage of training the RL system for SP localization (about 3 days), we retrained the agents and the RNN module with fixed architectures for 100 epochs using a learning rate of 5e-5. The batch sizes for training the uterus and fetal brain datasets are set as 32 and 24, respectively. The size of the replay buffer in DDQN is set as 15000 and the target network copies the parameters of the current network every 1500 steps. During training, the initialized planes were randomly set around the target planes within an angle and distance range of $\pm20^{\circ}$/ $\pm4mm$ and $\pm25^{\circ}$/ $\pm5mm$ for the uterus and fetal brain, respectively. The step sizes in each update iteration for angles (\emph{a}$_{\zeta}$, \emph{a}$_{\beta}$ and \emph{a}$_{\phi}$) are set as $\pm0.5^{\circ}$ and $\pm1.0^{\circ}$ for uterus and fetal brain, and the step size of distance \emph{a}$_{d}$ is set as $\pm0.1 mm$ for both. Besides, the termination steps are set as 50, 30 for the uterus and 80, 60 for the fetal brain in the training and testing phases, respectively.
	

	Extensive experiments were conducted to compare the proposed framework with the classical supervised learning (SL) methods and the state-of-the-art RL methods.
	The registration results obtained by transformation using the Atlas-based registration~\citep{dou2019agent} are considered as the baseline localization accuracy. 
	We first compared the SL-based methods, i.e., single-plane regression (S-Regression) and multi-plane regression (M-Regression), which take the whole 3D volume as input and output the regressed plane parameters ($\zeta,\beta,\phi, d$ in plane function $\cos(\zeta)x + \cos(\beta)y + \cos(\phi)z + d = 0$). 
	RL-based methods are also compared, including SARL and MARL.
	To further validate each component in the proposed framework, we also conduct ablation studies. In specific, to validate the effectiveness of the landmark-alignment module, we reported the results when the module is not applied, i.e., SARL without alignment module (SARL-WA). To test the contribution of the proposed RNN-based collaborative module, we implemented the MARL method without NAS while equipped it with the proposed RNN (MARL-R). To test whether NAS could further help the MARL and compare the performance of different NAS measures, we added DARTS and GDAS to the MARL framework (D-MARL and G-MARL). Furthermore, to validate the compound effect of the NAS and the RNN collaborative module, we implemented D-MARL-R and G-MARL-R. Note that the RNN modules in MARL-R, D-MARL-R, and G-MARL-R are classical hand-crafted BiLSTM~\citep{graves2005framewise}. We used 3D ResNet18 as backbone for S-Regression and M-Regression, and 2D ResNet18 served as the network backbone for SARL-WA, SARL, MARL and MARL-R.
	
	%
	%
	%
	
	\begin{table*}[!t]
		\scriptsize
		
		\caption{\centering{Localization results of SL and RL-based methods (batch size = 1 for fair comparion). The best results of S-Regression vs. SARL, and M-Regression vs. MARL are shown in blue.}}\label{regression1}
		
		\begin{center}
			\setlength{\tabcolsep}{2mm}{
				\begin{tabular}{c|c|c|c|c|c|c|c}
					\toprule
					& \diagbox [width=5em,trim=l]{Metrics} {Plane} & S & T & C & TT & TV & TC \\
					
					\hline
					&Ang ($^{\circ}$) &{11.55$\pm$9.99} & {9.96$\pm$7.18}&  {8.50$\pm$6.73}&{19.75$\pm$13.83}&{17.52$\pm$13.46}&{18.25$\pm$16.65}\\
					
					Registraion & Dis (mm)&{2.86$\pm$3.39}&{3.29$\pm$2.60}&{1.54$\pm$1.60}&{1.41$\pm$1.31} &{2.69$\pm$2.27}& {2.22$\pm$2.30}\\
					
					& SSIM &{0.81$\pm$0.08} & {0.67$\pm$0.10}&{0.54$\pm$0.12}&{0.68$\pm$0.11}&{0.66$\pm$0.10}&{0.68$\pm$0.11}\\
					
					\hline
					\hline
					&Ang ($^{\circ}$) &{10.44$\pm$8.28} & {9.88$\pm$9.72}&  {9.53$\pm$5.32}&{16.63$\pm$15.77}&{15.92$\pm$18.21}&{15.33$\pm$13.35}\\
					
					S-Regression & Dis (mm)&{3.17$\pm$3.11}&{3.23$\pm$2.29}&{2.32$\pm$1.30}&{2.33$\pm$1.44} &{3.55$\pm$2.19}& \textcolor{blue}{2.31$\pm$1.72}\\
					
					& SSIM &{0.83$\pm$0.12} & {0.70$\pm$0.11}&{0.52$\pm$0.10}&{0.71$\pm$0.12}&{0.74$\pm$0.11}&{0.78$\pm$0.12}\\

					\hline
					&Ang ($^{\circ}$) & \textcolor{blue}{9.92$\pm$9.71} & \textcolor{blue}{9.21$\pm$8.42}&  \textcolor{blue}{8.27$\pm$6.81}&\textcolor{blue}{16.23$\pm$14.31}&\textcolor{blue}{15.32$\pm$14.99}&\textcolor{blue}{14.77$\pm$15.21}\\
					
					SARL & Dis (mm)&\textcolor{blue}{2.92$\pm$3.31}&\textcolor{blue}{3.11$\pm$2.45}&\textcolor{blue}{1.55$\pm$1.31}&\textcolor{blue}{1.49$\pm$1.28} &\textcolor{blue}{3.47$\pm$2.34}& {2.42$\pm$1.96}\\
					
					& SSIM &\textcolor{blue}{0.87$\pm$0.08} & \textcolor{blue}{0.74$\pm$0.09}&\textcolor{blue}{0.69$\pm$0.09}&\textcolor{blue}{0.76$\pm$0.08}&\textcolor{blue}{0.75$\pm$0.06}&\textcolor{blue}{0.80$\pm$0.07}\\
					
					\hline
					\hline
					&Ang ($^{\circ}$) &{10.67$\pm$10.55} & {9.64$\pm$9.76}&  {9.37$\pm$8.90}&{16.77$\pm$14.21}&{15.89$\pm$16.02}&{15.21$\pm$14.99}\\
					
					M-Regression & Dis (mm)&{3.11$\pm$2.84}&{3.18$\pm$2.66}&{2.00$\pm$1.81}&{2.24$\pm$1.82} &{3.15$\pm$2.44}& {2.19$\pm$1.77}\\
					
					& SSIM &{0.85$\pm$0.07} & {0.73$\pm$0.09}&{0.57$\pm$0.08}&{0.72$\pm$0.10}&{0.75$\pm$0.09}&{0.79$\pm$0.11}\\
					
					\hline
					&Ang ($^{\circ}$) &\textcolor{blue}{9.88$\pm$9.47} & \textcolor{blue}{9.44$\pm$9.22}&  \textcolor{blue}{7.89$\pm$7.23}&\textcolor{blue}{16.08$\pm$14.55}&\textcolor{blue}{13.66$\pm$12.58}&\textcolor{blue}{13.02$\pm$14.79}\\
					
					MARL & Dis (mm)&\textcolor{blue}{2.44$\pm$2.11}&\textcolor{blue}{3.00$\pm$2.28}&\textcolor{blue}{1.38$\pm$1.72}&\textcolor{blue}{1.41$\pm$1.55} &\textcolor{blue}{1.55$\pm$1.42}& \textcolor{blue}{1.88$\pm$1.99}\\
					
					& SSIM &\textcolor{blue}{0.88$\pm$0.09} & \textcolor{blue}{0.75$\pm$0.11}&\textcolor{blue}{0.70$\pm$0.09}&\textcolor{blue}{0.75$\pm$0.08}&\textcolor{blue}{0.79$\pm$0.07}&\textcolor{blue}{0.81$\pm$0.08}\\
					\bottomrule
			\end{tabular}}
			\vspace{-0.6cm}
		\end{center}
	\end{table*}

	\begin{table*}[!t]
		\caption{\centering{P-values of paired t-test between the results of SL- and RL-based methods for all the evaluation metrics used in our study (Ang, Dis and SSIM). P-values below the significance level (0.05) are shown in blue.}}\label{ttest}
		\scriptsize
		\begin{center}
			\setlength{\tabcolsep}{2mm}{
				\begin{tabular}{c|c|c|c|c|c|c|c}
					\toprule
					& \diagbox [width=5em,trim=l]{Metrics} {Plane} & S & T & C & TT & TV & TC \\
					\hline
					&Ang ($^{\circ}$) &\textcolor{blue}{0.001} & \textcolor{blue}{0.021}&  \textcolor{blue}{10$^{-4}$}&{0.083}&\textcolor{blue}{0.026}&\textcolor{blue}{0.011}\\
					S-Regression vs. SARL& Dis (mm)&\textcolor{blue}{0.002}&{0.257}&\textcolor{blue}{10$^{-4}$}&\textcolor{blue}{10$^{-5}$} &{0.312}& {0.487}\\
					& SSIM &\textcolor{blue}{10$^{-4}$} & \textcolor{blue}{0.007}&\textcolor{blue}{10$^{-7}$}&\textcolor{blue}{10$^{-4}$}&{0.133}&{0.384}\\
					\hline
					\hline
					&Ang ($^{\circ}$) &\textcolor{blue}{0.006} & {0.182}&  \textcolor{blue}{10$^{-6}$}&\textcolor{blue}{0.001}&\textcolor{blue}{10$^{-7}$}&\textcolor{blue}{10$^{-7}$}\\
					M-Regression vs. MARL & Dis (mm)&\textcolor{blue}{0.001}&\textcolor{blue}{0.007}&\textcolor{blue}{0.002}&\textcolor{blue}{10$^{-4}$} &\textcolor{blue}{10$^{-7}$}& \textcolor{blue}{0.001}\\
					& SSIM &\textcolor{blue}{0.011} & {0.172}&\textcolor{blue}{10$^{-5}$}& {0.124}&\textcolor{blue}{0.017}&{0.133}\\
					\bottomrule
			\end{tabular}}
			\vspace{-0.45cm}
		\end{center}
	\end{table*}
	
	\begin{table*}[!t]
		\scriptsize
		\caption{Quantitative evaluation of plane localization in uterus US volume. The best results are shown in blue.}\label{u_table}
		\begin{center}
			\setlength{\tabcolsep}{2.5mm}{
				\begin{tabular}{c|c|c|c|c|c|c|c|c|c|c}
					\toprule
					& Metrics & SARL-WA & SARL & MARL & MARL-R & D-MARL & D-MARL-R & G-MARL & G-MARL-R & Ours \\
					\hline
					&Ang ($^{\circ}$) &{44.12$\pm$14.00} & {9.68$\pm$9.63}&  {10.37$\pm$10.18}&{8.48$\pm$9.31}&{8.71$\pm$8.66}&{7.88$\pm$9.32}&{8.66$\pm$8.24}&{7.09$\pm$8.47} &\textcolor{blue}{6.88$\pm$8.11}\\
					S & Dis (mm)&{15.27$\pm$9.87}&{2.84$\pm$3.49}&{2.00$\pm$2.38}&{2.41$\pm$3.19} &{2.17$\pm$3.27}&{2.08$\pm$2.81}&{2.19$\pm$2.97} & {2.15$\pm$3.07} &\textcolor{blue}{1.57$\pm$2.73}\\
					& SSIM &{0.74$\pm$0.07} & {0.88$\pm$0.06}&{0.89$\pm$0.09}&{0.89$\pm$0.07}&{0.88$\pm$0.06}&{0.89$\pm$0.06}&{0.87$\pm$0.13}&{0.90$\pm$0.06} &\textcolor{blue}{0.91$\pm$0.07}\\
					
					\hline
					& Ang ($^{\circ}$)& {66.71$\pm$17.31}& {9.53$\pm$8.27}&{9.30$\pm$8.87}&{8.87$\pm$7.37}&{8.91$\pm$7.55}&{8.67$\pm$7.67}&{8.69$\pm$8.77}&{8.35$\pm$8.67} &\textcolor{blue}{8.30$\pm$7.96}\\
					T & Dis (mm) &{16.48$\pm$13.02} & {3.17$\pm$2.58}&{2.99$\pm$2.61}&{2.69$\pm$2.49}&{2.28$\pm$2.20}&{2.31$\pm$2.67}&{2.21$\pm$2.08}&\textcolor{blue}{2.11$\pm$2.34} &{2.26$\pm$2.71}\\
					& SSIM & {0.58$\pm$0.07} &{0.75$\pm$0.10}&{0.72$\pm$0.10}&{0.71$\pm$0.11}&{0.74$\pm$0.11}&{0.75$\pm$0.12}&{0.75$\pm$0.11}&\textcolor{blue}{0.76$\pm$0.12}&{0.75$\pm$0.11}\\
					
					\hline
					& Ang ($^{\circ}$) & {53.21$\pm$13.21}& {8.00$\pm$6.76}&{7.14$\pm$6.67}&{7.17$\pm$6.13}&{7.14$\pm$1.22}&{6.31$\pm$8.02}&{7.23$\pm$6.51}&{6.40$\pm$6.42}&\textcolor{blue}{5.92$\pm$6.32}\\
					C & Dis (mm) & {13.41$\pm$14.47}& {1.46$\pm$1.40}&{1.53$\pm$1.58}&{1.22$\pm$1.27}&{1.33$\pm$1.48}&{0.95$\pm$1.13}&{1.20$\pm$1.34}&\textcolor{blue}{0.83$\pm$1.30}&{0.93$\pm$1.29}\\
					& SSIM &{0.52$\pm$0.06} &{0.69$\pm$0.09}&{0.67$\pm$0.09}&{0.68$\pm$0.09}&{0.68$\pm$0.10}&{0.73$\pm$0.13}&{0.68$\pm$0.11}&{0.73$\pm$0.11}&\textcolor{blue}{0.75$\pm$0.11}\\
					
					\hline
					& Ang ($^{\circ}$) &{54.68$\pm$15.23} & {9.07$\pm$8.34}&{8.94$\pm$8.79}&{8.17$\pm$8.13}&{8.25$\pm$7.89}&{7.62$\pm$8.04}&{8.19$\pm$8.26}&{7.28$\pm$8.15}&\textcolor{blue}{7.03$\pm$6.92}\\
					Avg & Dis (mm) &{15.05$\pm$12.87}& {2.49$\pm$2.74}&{2.17$\pm$2.31}&{2.11$\pm$2.53}&{1.93$\pm$2.41}&{1.78$\pm$2.11}&{1.87$\pm$2.40}&{1.70$\pm$2.37}&\textcolor{blue}{1.59$\pm$2.48}\\
					& SSIM &{0.61$\pm$0.13}& {0.77$\pm$0.12}&{0.76$\pm$0.13}&{0.76$\pm$0.13}&{0.77$\pm$0.13}&{0.79$\pm$0.13}&{0.77$\pm$0.12}&{0.80$\pm$0.13}&\textcolor{blue}{0.81$\pm$0.12} \\
					\bottomrule
			\end{tabular}}
			\vspace{-0.45cm}
		\end{center}
	\end{table*}
	
	\begin{table*}[!t]
		\scriptsize
		\caption{Quantitative evaluation of plane localization in fetal brain US volume. The best results are shown in blue.}\label{f_table}
		\begin{center}
			\setlength{\tabcolsep}{2mm}{
				\begin{tabular}{c|c|c|c|c|c|c|c|c|c|c}
					\toprule
					& Metrics & SARL-WA & SARL & MARL & MARL-R & D-MARL & D-MARL-R & G-MARL & G-MARL-R & Ours \\
					\hline
					&Ang ($^{\circ}$) &{57.35$\pm$12.31} & {16.11$\pm$13.42}&  {15.94$\pm$11.01}&{13.60$\pm$15.30}&{13.34$\pm$12.11}&{11.76$\pm$13.71}&{12.51$\pm$14.14}&{11.34$\pm$13.11} &\textcolor{blue}{11.06$\pm$11.77}\\
					TT & Dis (mm)&{9.92$\pm$6.29}&{1.37$\pm$1.27}&{1.34$\pm$1.27}&{1.21$\pm$1.03} &{1.32$\pm$1.24}&\textcolor{blue}{0.92$\pm$1.01}&{1.32$\pm$0.89} & {0.93$\pm$1.06} &{0.94$\pm$1.01}\\
					& SSIM &{0.55$\pm$0.07} & {0.77$\pm$0.10}&{0.73$\pm$0.10}&{0.74$\pm$0.18}&{0.77$\pm$0.12}&{0.78$\pm$0.12}&{0.78$\pm$0.14}&{0.82$\pm$0.12} &\textcolor{blue}{0.83$\pm$0.11}\\
					\hline
					
					& Ang ($^{\circ}$)& {48.42$\pm$12.45}& {15.27$\pm$13.91}&{9.75$\pm$11.86}&{9.72$\pm$14.00}&{9.83$\pm$10.61}&{9.61$\pm$11.48}&{9.89$\pm$12.56}&{9.54$\pm$12.89} &\textcolor{blue}{8.59$\pm$12.51}\\
					TV& Dis (mm) &{14.64$\pm$10.20} & {3.40$\pm$1.97}&{1.68$\pm$2.10}&\textcolor{blue}{1.30$\pm$1.74}&{1.72$\pm$2.05}&{1.46$\pm$1.17}&{1.66$\pm$1.93}&{1.50$\pm$2.13} &{1.43$\pm$1.98}\\
					& SSIM & {0.59$\pm$0.05} &{0.76$\pm$0.09}&{0.80$\pm$0.10}&{0.80$\pm$0.14}&{0.79$\pm$0.12}&{0.80$\pm$0.12}&{0.79$\pm$0.11}&{0.83$\pm$0.11}&\textcolor{blue}{0.84$\pm$0.11}\\
					
					\hline
					& Ang ($^{\circ}$) & {48.31$\pm$18.25}& {14.43$\pm$17.32}&{12.23$\pm$18.03}&{12.48$\pm$17.90}&{11.73$\pm$15.35}&{11.02$\pm$14.79}&{11.26$\pm$15.66}&{10.04$\pm$15.03}&\textcolor{blue}{9.62$\pm$14.34}\\
					TC & Dis (mm) & {10.55$\pm$7.46}& {2.20$\pm$2.00}&{1.62$\pm$2.19}&{1.25$\pm$1.74}&{1.49$\pm$1.51}&{1.23$\pm$1.09}&{1.41$\pm$1.89}&{1.27$\pm$1.98}&\textcolor{blue}{1.20$\pm$1.76}\\
					& SSIM &{0.58$\pm$0.05} &{0.80$\pm$0.08}&{0.78$\pm$0.10}&{0.81$\pm$0.12}&{0.79$\pm$0.09}&{0.80$\pm$0.11}&{0.80$\pm$0.10}&{0.83$\pm$0.11}&\textcolor{blue}{0.84$\pm$0.10}\\
					
					\hline
					& Ang ($^{\circ}$) &{51.36$\pm$15.06} & {15.27$\pm$15.11}&{12.64$\pm$14.22}&{11.93$\pm$15.90}&{11.63$\pm$14.67}&{10.80$\pm$13.91}&{11.22$\pm$14.22}&{10.31$\pm$13.74}&\textcolor{blue}{9.75$\pm$12.92}\\
					Avg & Dis (mm) &{11.70$\pm$6.57}& {2.32$\pm$1.97}&{1.55$\pm$1.91}&{1.25$\pm$1.55}&{1.51$\pm$1.44}&{1.20$\pm$1.43}&{1.32$\pm$1.67}&{1.23$\pm$1.81}&\textcolor{blue}{1.19$\pm$1.66}\\
					& SSIM &{0.57$\pm$0.06}& {0.78$\pm$0.09}&{0.77$\pm$0.12}&{0.78$\pm$0.15}&{0.78$\pm$0.13}&{0.79$\pm$0.13}&{0.79$\pm$0.12}&{0.83$\pm$0.10}&\textcolor{blue}{0.84$\pm$0.11} \\
					\bottomrule
			\end{tabular}}
			\vspace{-0.45cm}
		\end{center}
	\end{table*}
	
	\begin{table}[!t]
		\caption{Quantitative evaluation of plane localization for normal and abnormal uterus.}\label{comparison_uterus_result}
		\begin{center}
			\scriptsize
			\begin{tabular}{c|c|c|c}
				\toprule
				& Avg-Ang ($^{\circ}$) & Avg-Dis (mm) & Avg-SSIM \\
				\hline
				\textbf{Normal} & {6.28$\pm$7.01} &  {1.57$\pm$1.53}&{0.83$\pm$0.07} \\
				\hline
				\textbf{Abnormal} & {8.87$\pm$7.97} & {1.63$\pm$1.26}&{0.74$\pm$0.08} \\
				\bottomrule
			\end{tabular}
			\vspace{-0.45cm}
		\end{center}
	\end{table}
	
	\begin{table*}[!h]
		\caption{Model information of compared methods.}\label{information}
		\begin{center}
			\scriptsize
			\begin{tabular}{c|c|c|c|c|c|c|c|c|c|c|c}
				\toprule
				& & S-Regression & M-Regression & SARL& MARL & MARL-R & D-MARL & D-MARL-R & G-MARL & G-MARL-R & Ours\\
				\hline
				\multirow{2}*{\textbf{FLOPs (G)}} & Uterus & \multirow{2}* {337.23 *3} & \multirow{2}* {337.26} & \multirow{2}* {1.82*3} &\multirow{2}* {1.82}& \multirow{2}* {1.82}&{0.69}&{0.69}&{0.68}&{0.68}&{0.68}\\
				\cline{2-2}
				\cline{8-12}
				
				& Fetal Brain & &&& &&{0.72}&{0.72}&{0.70}&{0.70}&{0.70}\\
				\hline
				\multirow{2}*{\textbf{Params (M)}} & Uterus &\multirow{2}* {33.16 *3} &\multirow{2}* {33.17}&\multirow{2}* {12.21*3}& \multirow{2}* {12.22}&\multirow{2}* {12.26}&{3.62}&{3.66}&{3.61}&{3.66}&{3.65}\\
				\cline{2-2}
				\cline{8-12}
				& Fetal Brain & & &&& &{4.69}&{4.73}&{4.52}&{4.56}&{4.56}\\
				\bottomrule
			\end{tabular}
			\vspace{-0.45cm}
		\end{center}
	\end{table*}
	
	\begin{table*}[!h]
		\scriptsize
		\caption{Scoring results of our proposed method.}\label{score}
		\begin{center}
			\setlength{\tabcolsep}{2mm}{
				\begin{tabular}{c|c|c|c|c|c|c}
					\toprule
					& S & T & C & TT & TV & TC \\
					\hline
					Score (Mean$\pm$STD) &{0.763$\pm$0.131}&{0.690$\pm$0.111}&{0.769$\pm$0.107}&{0.803$\pm$0.130} &{0.759$\pm$0.124}& {0.783$\pm$0.133}\\
					\hline
					Qualified rate (Score $>$0.6) &{93.1\%}&{86.1\%}&{95.1\%}&{97.05\%} &{96.08\%}& {96.08\%}\\
					
					\bottomrule
			\end{tabular}}
			\vspace{-0.45cm}
		\end{center}
	\end{table*}

	\subsection{Quantitative Results}
	
	Tab.~\ref{regression1} shows the results of our baseline method (i.e., Registration), SL- and RL-based methods (i.e. S-Regression vs. SARL and M-Regression vs. MARL).
	Note that for a fair comparison, we carefully designed the network structure (3D\&2D ResNet18) and training strategy (Adam with learning rate = 5e-5 and batch size = 1) for SL- and RL-based methods. 
	It can be seen that the RL-based methods can achieve better performance than the SL-based algorithms in almost all the evaluation metrics of different planes.
	To further investigate the statistically significant difference, we performed paired t-tests (significance level = 0.05) between the results of S-Regression vs. SARL and M-Regression vs. MARL in Tab.~\ref{ttest}. 
	It shows that the differences in most evaluation metrics between the two methods have statistical significance.
	This proves that directly learning the mapping from high-dimension volumetric data to low-dimension abstract features (i.e., plane parameters) is more difficult than learning that from 2D plane data in a sonographers-like RL system (refers to Sec.~\ref{Sec:RL}). Thus, RL-based methods are more reasonable for our task of plane localization in 3D US.
	Tab.~\ref{u_table} and Tab.~\ref{f_table} report the RL results of the uterus and the fetal brain datasets, respectively. 
	The results of `SARL-WA'  and `SARL' reflect that without the landmark-alignment module, the agents are difficult to train and may fail to find their targets due to the complex initial environment, thus causing the large localization errors.
	Compare the `SARL' with the `MARL' column, it can be seen that the MARL-based methods achieved better performance than the SARL. This validates previous conjecture that multiple agents can share their useful information and improve the performance when detecting multiple SPs. Considering the necessity of NAS, both the `D-MARL' and the `G-MARL' outperformed plain `MARL' in both scenarios. Furthermore, `G-MARL' scored higher accuracy than `D-MARL', while searching the network architecture faster than the latter one (refers to Section.~\ref{Qualitative results}). Another interesting phenomenon is that the RNN-based collaborative module is able to boost the model performance consistently (i.e. MARL-R, D-MARL-R and G-MARL-R). It proves that interaction and communication among different tasks are beneficial, as it provides knowledge on the spatial relationship between SPs. The last column of Tab.~\ref{u_table} and Tab.~\ref{f_table} shows the accuracy of the proposed framework, it can be seen that it outperformed its counterparts in almost all metrics in both datasets.

	Another point that should be discussed is whether the proposed framework can handle both the normal and the abnormal cases of the uterus. Tab.~\ref{comparison_uterus_result} reports the performance of our system in detecting SPs of normal and abnormal cases. Our system achieved comparable localization accuracies in terms of orientation and position of the target SPs in both cases. It proves that the proposed methods can be applied to real clinical scenarios where challenging abnormal cases present.

	Additionally, the computational cost and the size of a model are also a matter of practical concern. Previous methods~\citep{alansary2018automatic,dong2019searching} adopted the VGG model as the backbone, which has been widely adopted in many computer vision tasks. In this work, we opt for ResNet18 which is also popular but relatively smaller, thus reducing both FLOPs and Params. Tab.~\ref{information} shows the FLOPs (computational costs) and the Params (model sizes) of all the tested methods. 
	First, it can be seen that `SARL' and `MARL' are more resource-saving than `S-Regression' and `M-Regression', respectively.
	Furthermore, compared with `SARL', `MARL' has fewer FLOPs and Params. Intuitively, the more planes are located, the more network parameters and the FLOPs will drop. Besides, the DARTS-based and GDAS-based methods (`D-MARL', `D-MARL-R', `G-MARL', `G-MARL-R' and `Ours') are even more lightweight compared to the hand-crafted networks (`SARL', `MARL' and `MARL-R'). The `G-MARL-R' model has comparable FLOPs and parameters with the proposed method. However, the proposed method achieved better performance in both datasets. It shows that the automatically discovered RNN is superior to the classical hand-crafted ones. It is interesting to point out that the automatically searched Neural Network models using the fetal brain dataset have more parameters than that of the one searched using the uterus dataset. We speculate that as the fetal brain has more anatomical structures and larger intra-class variations caused by brain development, it might require more complex network models to detect SPs in fetal brains. This finding also highlights the significance of using NAS to discover suitable neural network models for different tasks, which is more robust and does not necessitate human intervention.
	
	\begin{figure*}[!t]
		\centering
		\includegraphics[width=0.85\linewidth]{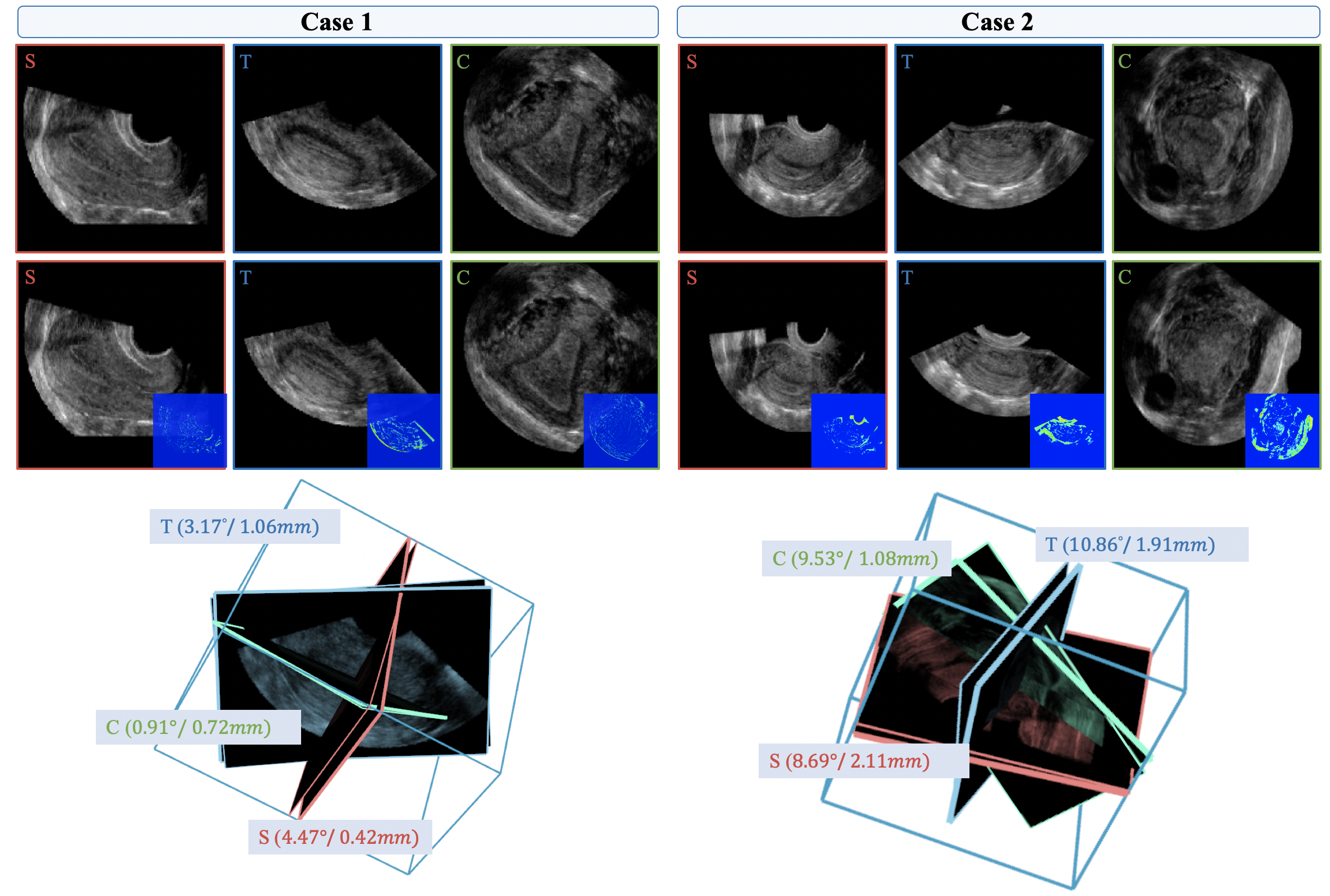}
		\caption{Visualization: plane localization results for normal uterus (left) and abnormal uterus with endometrium disease (right). In both two cases, the ground truth and prediction are shown in the first and second rows respectively (the pseudo-color images in the lower right corner visualize the absolute intensity errors between the ground truth and prediction), and the third row illustrates their spatial difference.}
		\label{fig:uterus}
		\vspace{-0.4cm}
	\end{figure*}
	
	\begin{figure*}[!h]
		\centering
		\includegraphics[width=0.85\linewidth]{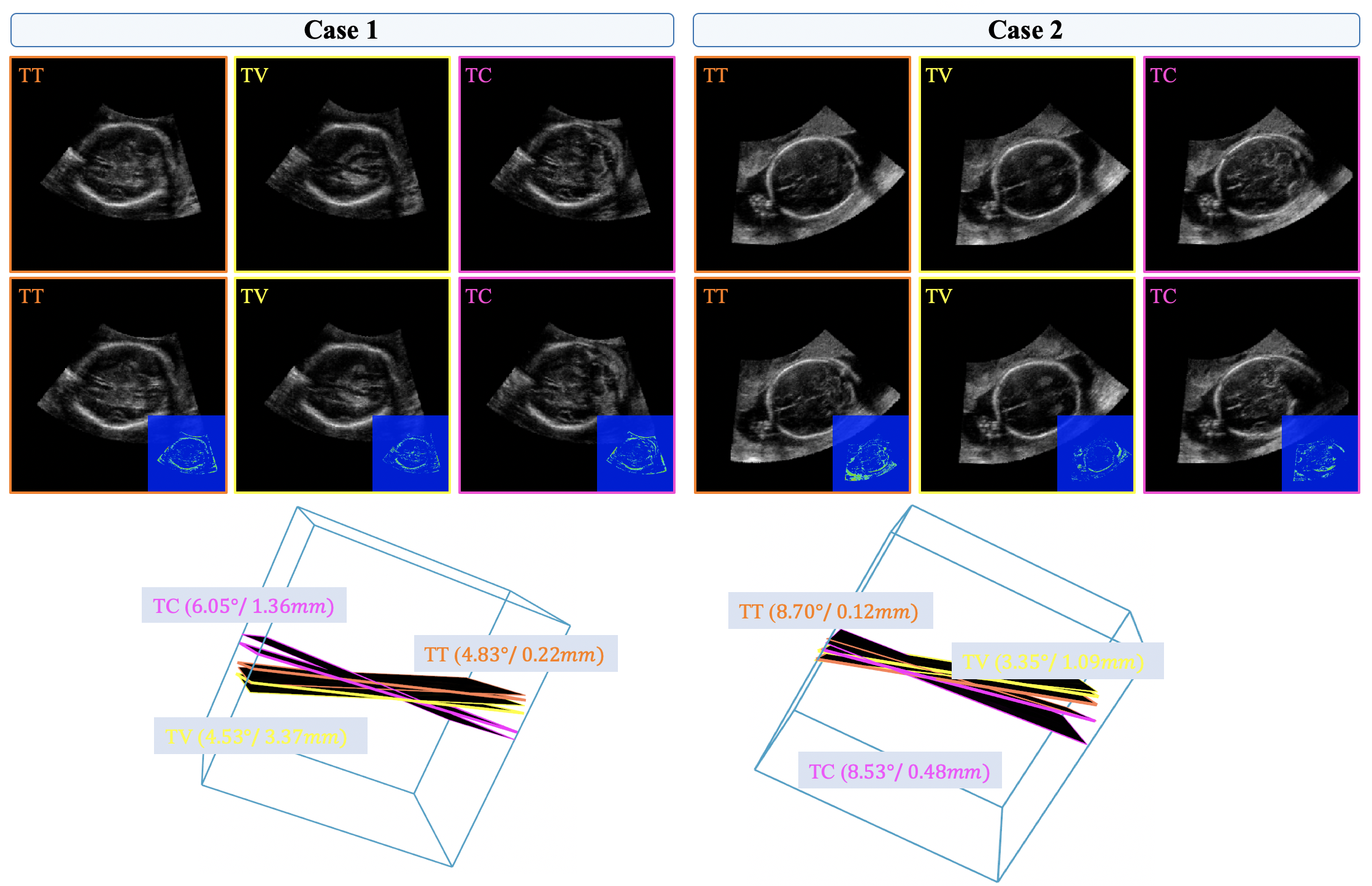}
		\caption{Visualization: plane localization results for two fetal brain volumes (left and right). In both two cases, the ground truth and prediction are shown in the first and second rows respectively (the pseudo-color images in the lower right corner visualize the absolute intensity errors between the ground truth and prediction), and the third row illustrates their spatial difference.}
		\label{fig:fetal_brain}
		\vspace{-0.6cm}
	\end{figure*}
	
	Furthermore, we performed the expert evaluation for the predicted planes from the clinical perspective. In specific, we invited four 5-year experienced sonographers to score the predicted planes based on 1) the existence of anatomical structures and 2) visual plausibility. 
	According to the clinical scoring standards, a plane with score$>$0.6 means that it contains the corresponding major anatomical structures and is acceptable in clinical diagnosis. 
	Thus we defined the score of a plane larger than 0.6 as $small$ $error$, which was determined by four experts based on the international clinical standards. 
	Note that four experts are divided into two groups (each group has two experts) to evaluate the uterus and fetal brain planes, respectively. The $score$ of one plane is calculated by averaging two experts' ratings, and the $qualified$ $rate$ is denoted as the percentage of planes with $score>$0.6 to the total. As shown in Tab.~\ref{score}, the localization results of our proposed method have small errors and can satisfy the clinical requirements well.

	
	\subsection{Qualitative Results}
	\label{Qualitative results}
	\begin{figure}[!t]
		\centering
		\includegraphics[width= 0.65\linewidth]{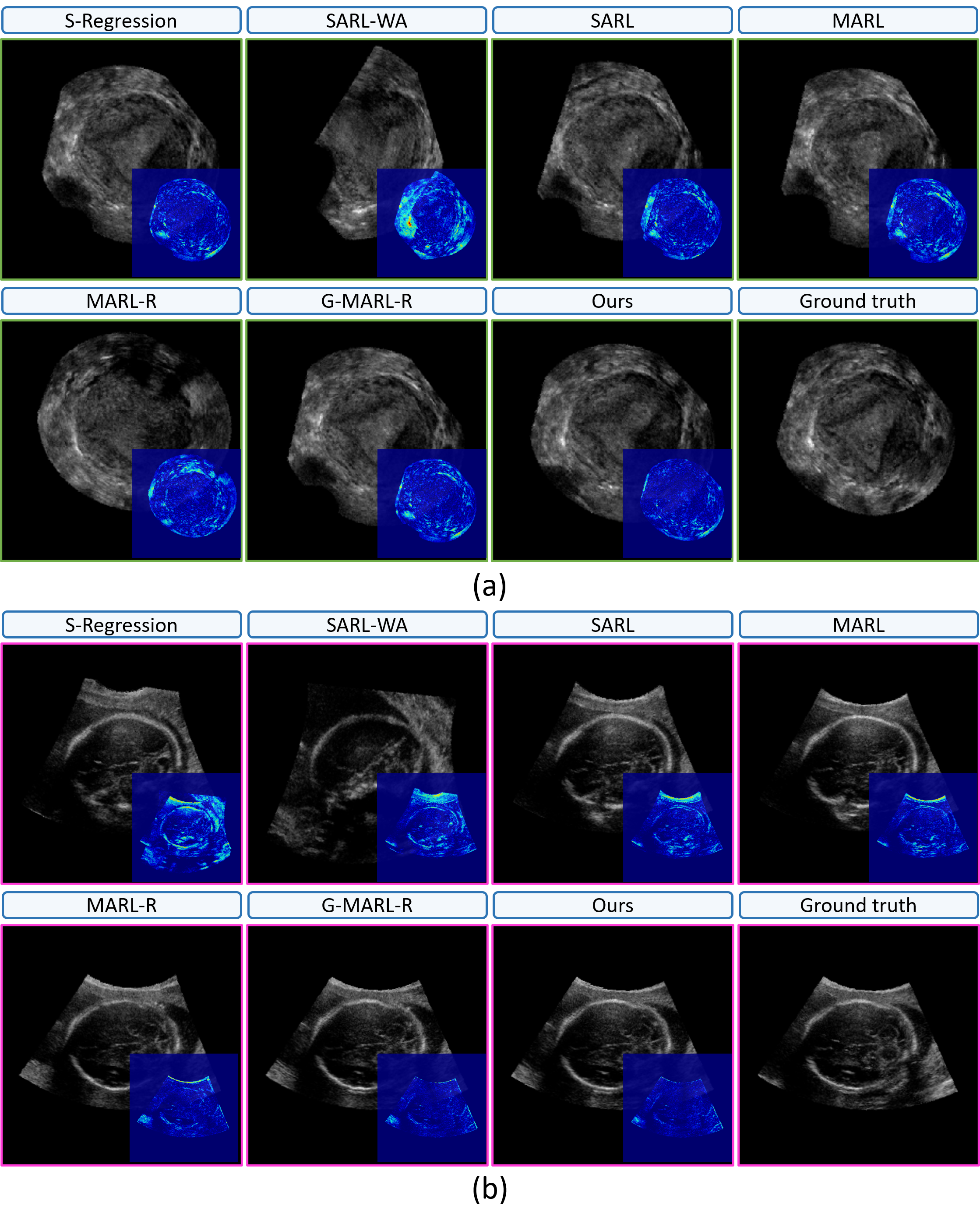}
		\caption{(a): C planes in uterus US volume and (b) TC planes in fetal brain US volume obtained by different methods. The pseudo-color images in the lower right corner visualize the absolute intensity errors between the ground truth and prediction.}\label{fig:cr}
		\vspace{-0.4cm}
	\end{figure}
	
	\begin{figure}[!htbp]
		\centering
		\includegraphics[width= 0.85\linewidth]{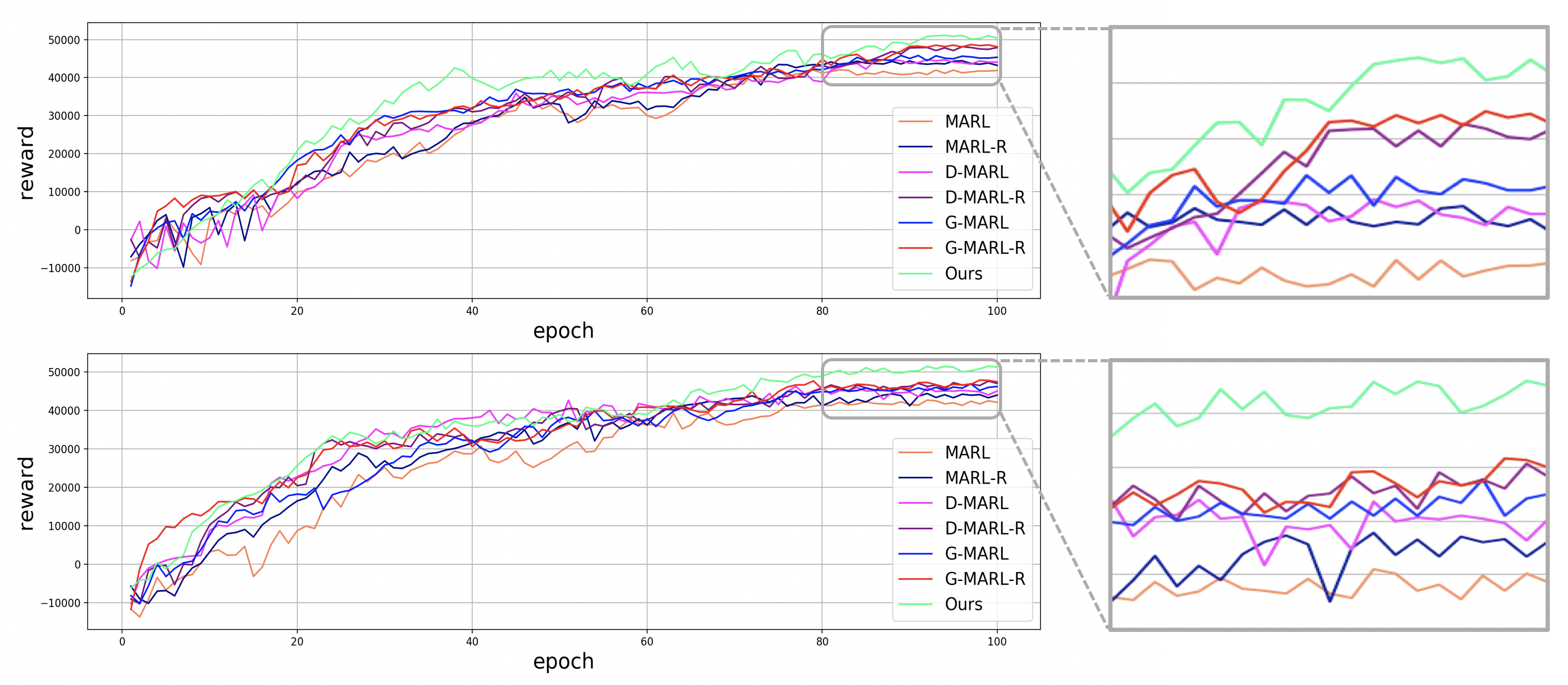}
		\caption{Comparison of accumulated reward curves from different methods during training for uterus ($top$) and fetal brain ($bottom$).}
		\label{fig:train_reward}
		\vspace{-0.05cm}
	\end{figure}
	Fig.~\ref{fig:uterus} and Fig.~\ref{fig:fetal_brain} provide visualization of predicted SPs for the uterine and fetal brain volumes, respectively. It shows that the detected planes are visually and spatially similar to the ground truth. It is interesting to see how the three SPs in fetal brain are closely adjacent. However, the proposed method is able to differentiate them and display related structures correctly. For example, the cerebellum (the `eye-glass' shape structure) can be seen clearly in the TC plane. Meanwhile, note that the second example in Fig.~\ref{fig:uterus} is especially challenging as this test case is diagnosed with endometrium disease. It can be seen that the endometrium is distorted (`C' plane) and is difficult to be recognized. However, the proposed method is still able to detect target SPs.
	Furthermore, in Fig.~\ref{fig:cr}, we show the C planes and TC planes obtained by our proposed methods ($Ours$) and the competing methods, including S-Regression, SARL-WA, SARL, MARL, MARL-R and G-MARL-R. From the pseudo-color images, it can be intuitively seen that planes predicted by our proposed method are closest to the ground truth.

	Fig.~\ref{fig:train_reward} shows the accumulated reward curves of MARL-related methods during \textit{training}. It is interesting to see that our method (green) can gain accumulated rewards higher and faster than its counterparts. Besides, `MARL-R' (dark blue), `D-MARL-R' (purple) and `G-MARL-R'(red) acquire higher accumulated rewards than `MARL' (orange), `D-MARL' (magenta) and `G-MARL'(blue) respectively, which indicates that the proposed RNN collaborative module can promote the communication among agents and thus achieving better performance. Meanwhile, `D-MARL' (magenta) and `G-MARL' (blue) also obtained higher accumulated rewards compare to `MARL' (orange) suggesting the effectiveness of incorporating NAS to search for a suitable architecture for each agent.

	\begin{figure*}[!t]
		\centering
		\includegraphics[width=0.9\linewidth]{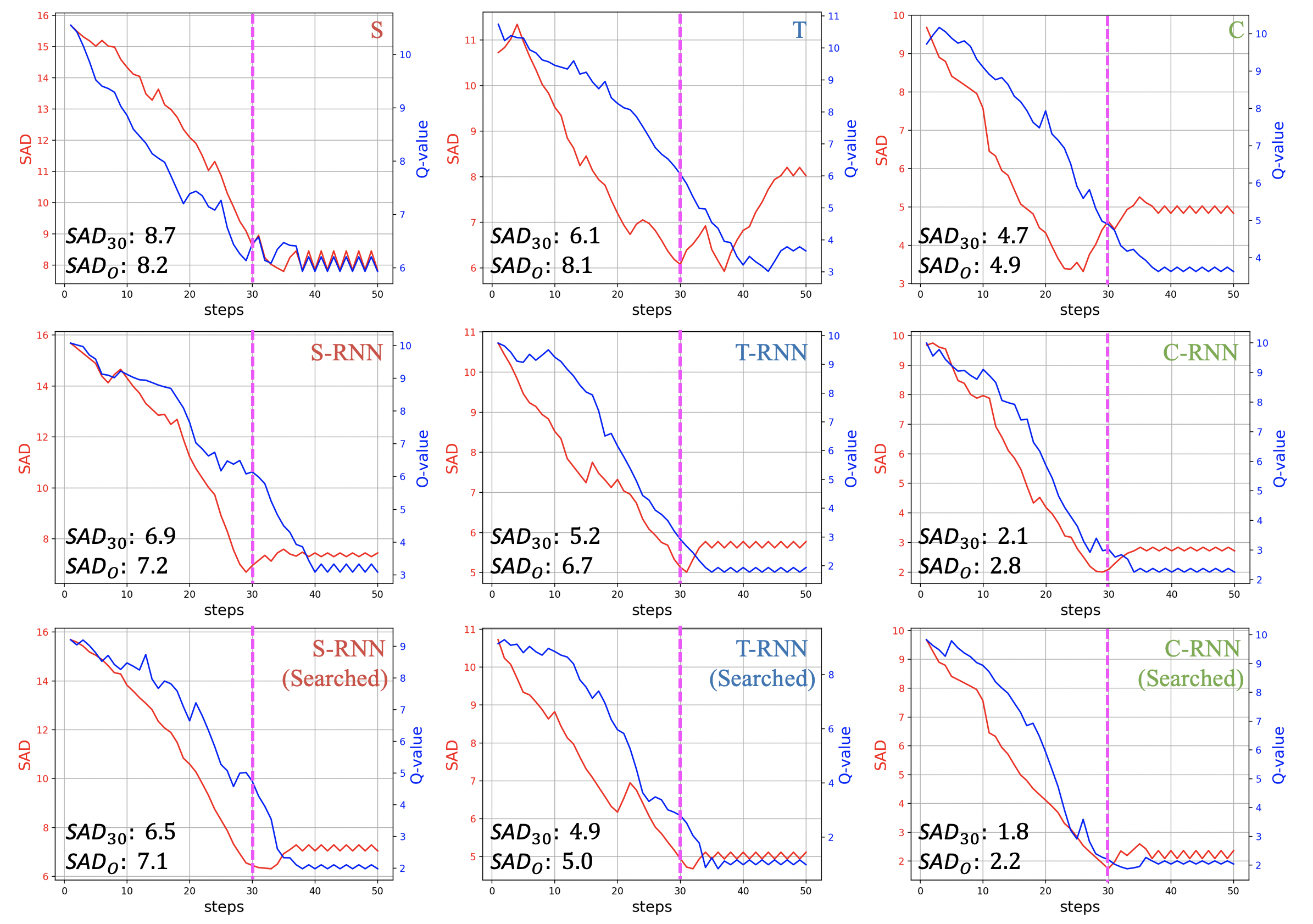}
		\centering
		\caption{SAD and Q-value curves of different agents in localizing three SPs (S, T and C) from one typical uterine US volume ($Top:$ without RNN, $Middle:$ with RNN (i.e. BiLSTM), $Bottom:$ with searched RNN). $SAD_{30}$ and $SAD_{o}$ calculate the SAD at step 30 (terminal step during testing, represented by magenta dotted lines) and the average SAD in the oscillating situation, respectively. Note that smaller $SAD_{*}$ values represent better localization performance.}
		\label{fig:q_value}
		\vspace{-0.4cm}
	\end{figure*}

	\begin{figure}[!t]
		\centering
		\includegraphics[width=0.7\linewidth]{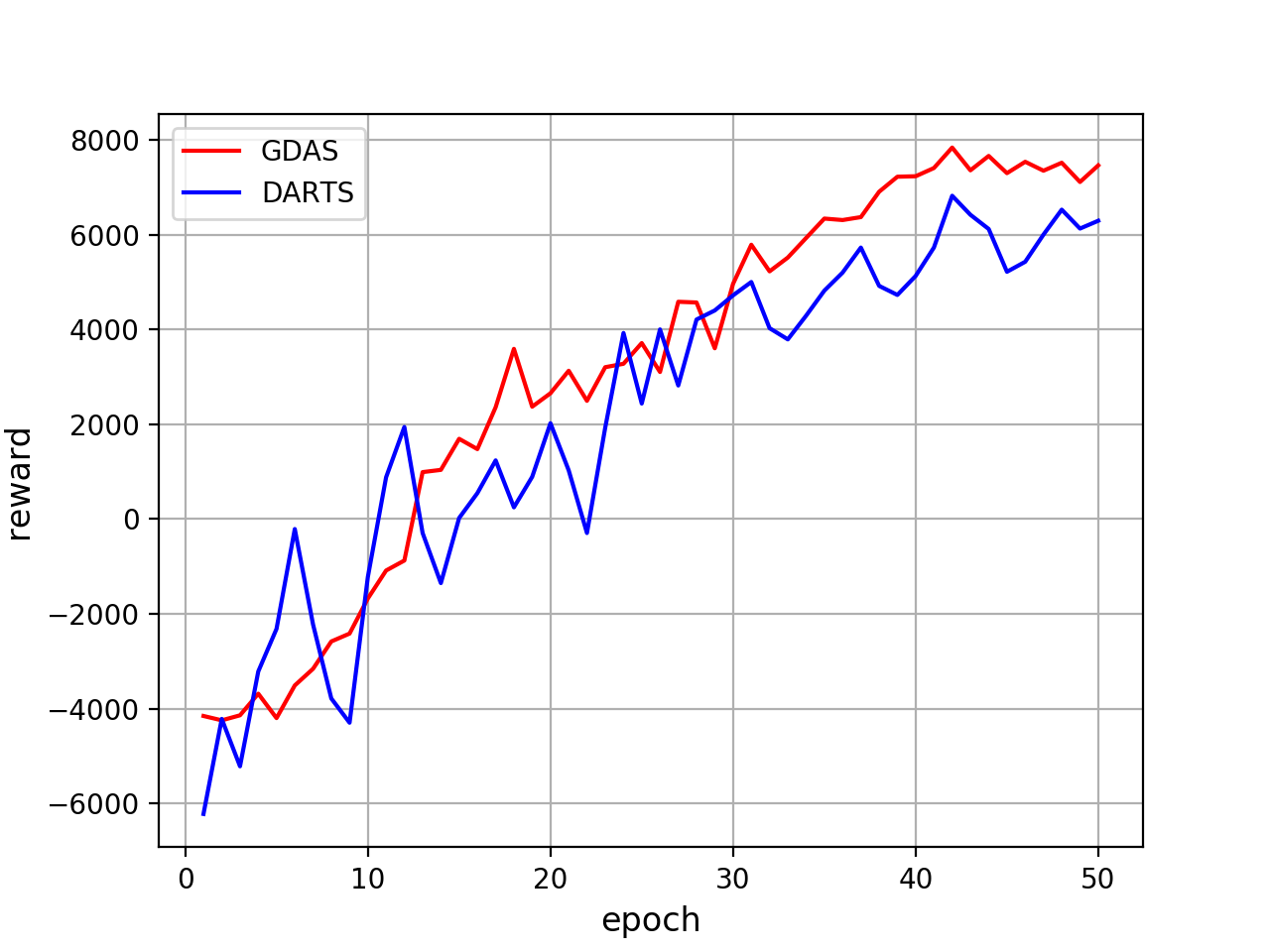}
		\centering
		\caption{Reward curve comparison between GDAS and DARTS during search. The DARTS curve is poor in stability and achieves a lower accumulated reward peak than the GDAS curve.}
		\label{fig:nas_compare}
		\vspace{-0.45cm}
	\end{figure}

	To validate the effectiveness of our proposed RNN based agent collaborative module deeply, we analyze the values of Q-value and the Sum of Angle and Distance (SAD). Specifically, the Q-value represents the quality of the action, and the SAD can reveal how close is the target to the ground truth. Fig.~\ref{fig:q_value} takes one typical uterine US volume as an example, and illustrates the SAD and Q-value curves of MARL models without RNN (`G-MARL'), with manual-designed RNN (`G-MARL-R') and with auto-searched RNN (`Ours'), respectively. It can be observed that the SAD and Q-value have a certain correlation (lower Q-value usually corresponds to lower SAD), while their final performances may oscillate differently (e.g. the last two sub-figures of `G-MARL'). However, the proposed collaborative module can enhance the learning of all agents, and make the SAD and Q-value converge to a better and more stable local minimum (see the last two rows of Fig.~\ref{fig:q_value}). Note that the SAD values (both $SAD_{30}$ and $SAD_{o}$) of `Ours' are lower than that of `G-MARL-R', it proves that our auto-searched RNN can be more effective to communicate the agents to improve the localization accuracy than the manual-designed RNN.

	Moreover, to compare the `D-MARL' and `G-MARL', it is also important to report their training time, occupied memory and stabilization during network searching. Experiments show that `G-MARL' (0.04 days per epoch, maximum batch size = 32) is much faster and memory-saving than `D-MARL' (0.12 days per epoch, maximum batch size = 4). Fig.~\ref{fig:nas_compare} shows the accumulated reward curves of `D-MARL' (blue) and `G-MARL' (red) during \textit{searching}. It can be seen that the blue curve oscillates more frequently than the red ones. This demonstrates that the searching process of GDAS is more stable than DARTS in our task.

	\section{Conclusion}
	\label{conclusion}
	In this paper, we developed a novel MARL framework to localize multiple SPs in 3D US. To incorporate the prior knowledge of the spatial relationship among planes, we proposed an RNN based agent collaborative module, which is general and can be utilized for other multi-task applications. Furthermore, the framework utilize GDAS to automatically search optimal agents for different SPs and also the architecture for the RNN collaborative module. The automatically discovered networks have better performance while having fewer parameters and FLOPs. Experimental results on two challenging 3D US datasets indicate that the proposed method is robust towards different clinical contexts and can localize the SPs effectively.
	
	\section*{Acknowledgments}
	This work was supported by the grant from National Key R$\&$D Program of China (No. 2019YFC0118300) and Shenzhen Peacock Plan (No. KQTD2016053112051497, KQJSCX20180328095606003).
	
	\bibliographystyle{model2-names.bst}\biboptions{authoryear}
	\bibliography{reference}

\begin{thebibliography}{48}
\expandafter\ifx\csname natexlab\endcsname\relax\def\natexlab#1{#1}\fi
\providecommand{\url}[1]{\texttt{#1}}
\providecommand{\href}[2]{#2}
\providecommand{\path}[1]{#1}
\providecommand{\DOIprefix}{doi:}
\providecommand{\ArXivprefix}{arXiv:}
\providecommand{\URLprefix}{URL: }
\providecommand{\Pubmedprefix}{pmid:}
\providecommand{\doi}[1]{\href{http://dx.doi.org/#1}{\path{#1}}}
\providecommand{\Pubmed}[1]{\href{pmid:#1}{\path{#1}}}
\providecommand{\bibinfo}[2]{#2}
\ifx\xfnm\relax \def\xfnm[#1]{\unskip,\space#1}\fi
\bibitem[{Alansary et~al.(2018)Alansary, Le~Folgoc, Vaillant, Oktay, Li, Bai,
  Passerat-Palmbach, Guerrero, Kamnitsas, Hou et~al.}]{alansary2018automatic}
\bibinfo{author}{Alansary, A.}, \bibinfo{author}{Le~Folgoc, L.},
  \bibinfo{author}{Vaillant, G.}, \bibinfo{author}{Oktay, O.},
  \bibinfo{author}{Li, Y.}, \bibinfo{author}{Bai, W.},
  \bibinfo{author}{Passerat-Palmbach, J.}, \bibinfo{author}{Guerrero, R.},
  \bibinfo{author}{Kamnitsas, K.}, \bibinfo{author}{Hou, B.}, et~al.,
  \bibinfo{year}{2018}.
\newblock \bibinfo{title}{Automatic view planning with multi-scale deep
  reinforcement learning agents}, in: \bibinfo{booktitle}{International
  Conference on Medical Image Computing and Computer-Assisted Intervention},
  \bibinfo{organization}{Springer}. pp. \bibinfo{pages}{277--285}.
\bibitem[{Baker et~al.(2016)Baker, Gupta, Naik and Raskar}]{baker2016designing}
\bibinfo{author}{Baker, B.}, \bibinfo{author}{Gupta, O.},
  \bibinfo{author}{Naik, N.}, \bibinfo{author}{Raskar, R.},
  \bibinfo{year}{2016}.
\newblock \bibinfo{title}{Designing neural network architectures using
  reinforcement learning}.
\newblock \bibinfo{journal}{arXiv preprint arXiv:1611.02167} .
\bibitem[{Baumgartner et~al.(2017)Baumgartner, Kamnitsas, Matthew, Fletcher,
  Smith, Koch, Kainz and Rueckert}]{Baumgartner2017SonoNetRD}
\bibinfo{author}{Baumgartner, C.F.}, \bibinfo{author}{Kamnitsas, K.},
  \bibinfo{author}{Matthew, J.}, \bibinfo{author}{Fletcher, T.P.},
  \bibinfo{author}{Smith, S.}, \bibinfo{author}{Koch, L.M.},
  \bibinfo{author}{Kainz, B.}, \bibinfo{author}{Rueckert, D.},
  \bibinfo{year}{2017}.
\newblock \bibinfo{title}{Sononet: Real-time detection and localisation of
  fetal standard scan planes in freehand ultrasound}.
\newblock \bibinfo{journal}{IEEE Transactions on Medical Imaging}
  \bibinfo{volume}{36}, \bibinfo{pages}{2204--2215}.
\bibitem[{Baumgartner et~al.(2016)Baumgartner, Kamnitsas, Matthew, Smith, Kainz
  and Rueckert}]{baumgartner2016real}
\bibinfo{author}{Baumgartner, C.F.}, \bibinfo{author}{Kamnitsas, K.},
  \bibinfo{author}{Matthew, J.}, \bibinfo{author}{Smith, S.},
  \bibinfo{author}{Kainz, B.}, \bibinfo{author}{Rueckert, D.},
  \bibinfo{year}{2016}.
\newblock \bibinfo{title}{Real-time standard scan plane detection and
  localisation in fetal ultrasound using fully convolutional neural networks},
  in: \bibinfo{booktitle}{International conference on medical image computing
  and computer-assisted intervention}, \bibinfo{organization}{Springer}. pp.
  \bibinfo{pages}{203--211}.
\bibitem[{Bornstein et~al.(2010)Bornstein, Monteagudo, Santos, Strock, Tsymbal,
  Lenchner and Timor-Tritsch}]{bornstein2010basic}
\bibinfo{author}{Bornstein, E.}, \bibinfo{author}{Monteagudo, A.},
  \bibinfo{author}{Santos, R.}, \bibinfo{author}{Strock, I.},
  \bibinfo{author}{Tsymbal, T.}, \bibinfo{author}{Lenchner, E.},
  \bibinfo{author}{Timor-Tritsch, I.}, \bibinfo{year}{2010}.
\newblock \bibinfo{title}{Basic as well as detailed neurosonograms can be
  performed by offline analysis of three-dimensional fetal brain volumes}.
\newblock \bibinfo{journal}{Ultrasound in obstetrics \& gynecology}
  \bibinfo{volume}{36}, \bibinfo{pages}{20--25}.
\bibitem[{Bu{\c{s}}oniu et~al.(2010)Bu{\c{s}}oniu, Babu{\v{s}}ka and
  De~Schutter}]{bucsoniu2010multi}
\bibinfo{author}{Bu{\c{s}}oniu, L.}, \bibinfo{author}{Babu{\v{s}}ka, R.},
  \bibinfo{author}{De~Schutter, B.}, \bibinfo{year}{2010}.
\newblock \bibinfo{title}{Multi-agent reinforcement learning: An overview}, in:
  \bibinfo{booktitle}{Innovations in multi-agent systems and applications-1}.
  \bibinfo{publisher}{Springer}, pp. \bibinfo{pages}{183--221}.
\bibitem[{Chen et~al.(2015)Chen, Ni, Qin, Li, Yang, Wang and
  Heng}]{chen2015standard}
\bibinfo{author}{Chen, H.}, \bibinfo{author}{Ni, D.}, \bibinfo{author}{Qin,
  J.}, \bibinfo{author}{Li, S.}, \bibinfo{author}{Yang, X.},
  \bibinfo{author}{Wang, T.}, \bibinfo{author}{Heng, P.A.},
  \bibinfo{year}{2015}.
\newblock \bibinfo{title}{Standard plane localization in fetal ultrasound via
  domain transferred deep neural networks}.
\newblock \bibinfo{journal}{IEEE journal of biomedical and health informatics}
  \bibinfo{volume}{19}, \bibinfo{pages}{1627--1636}.
\bibitem[{Chen et~al.(2017)Chen, Wu, Dou, Qin, Li, Cheng, Ni and
  Heng}]{chen2017ultrasound}
\bibinfo{author}{Chen, H.}, \bibinfo{author}{Wu, L.}, \bibinfo{author}{Dou,
  Q.}, \bibinfo{author}{Qin, J.}, \bibinfo{author}{Li, S.},
  \bibinfo{author}{Cheng, J.Z.}, \bibinfo{author}{Ni, D.},
  \bibinfo{author}{Heng, P.A.}, \bibinfo{year}{2017}.
\newblock \bibinfo{title}{Ultrasound standard plane detection using a composite
  neural network framework}.
\newblock \bibinfo{journal}{IEEE transactions on cybernetics}
  \bibinfo{volume}{47}, \bibinfo{pages}{1576--1586}.
\bibitem[{Chykeyuk et~al.(2013)Chykeyuk, Yaqub and Noble}]{chykeyuk2013class}
\bibinfo{author}{Chykeyuk, K.}, \bibinfo{author}{Yaqub, M.},
  \bibinfo{author}{Noble, J.A.}, \bibinfo{year}{2013}.
\newblock \bibinfo{title}{Class-specific regression random forest for accurate
  extraction of standard planes from 3d echocardiography}, in:
  \bibinfo{booktitle}{International MICCAI Workshop on Medical Computer
  Vision}, \bibinfo{organization}{Springer}. pp. \bibinfo{pages}{53--62}.
\bibitem[{Dong and Yang(2019)}]{dong2019searching}
\bibinfo{author}{Dong, X.}, \bibinfo{author}{Yang, Y.}, \bibinfo{year}{2019}.
\newblock \bibinfo{title}{Searching for a robust neural architecture in four
  gpu hours}, in: \bibinfo{booktitle}{Proceedings of the IEEE Conference on
  Computer Vision and Pattern Recognition}, pp. \bibinfo{pages}{1761--1770}.
\bibitem[{Dou et~al.(2019)Dou, Yang, Qian, Xue, Qin, Wang, Yu, Wang, Xiong,
  Heng et~al.}]{dou2019agent}
\bibinfo{author}{Dou, H.}, \bibinfo{author}{Yang, X.}, \bibinfo{author}{Qian,
  J.}, \bibinfo{author}{Xue, W.}, \bibinfo{author}{Qin, H.},
  \bibinfo{author}{Wang, X.}, \bibinfo{author}{Yu, L.}, \bibinfo{author}{Wang,
  S.}, \bibinfo{author}{Xiong, Y.}, \bibinfo{author}{Heng, P.A.}, et~al.,
  \bibinfo{year}{2019}.
\newblock \bibinfo{title}{Agent with warm start and active termination for
  plane localization in 3d ultrasound}, in: \bibinfo{booktitle}{International
  Conference on Medical Image Computing and Computer-Assisted Intervention},
  \bibinfo{organization}{Springer}. pp. \bibinfo{pages}{290--298}.
\bibitem[{Elsken et~al.(2018)Elsken, Metzen and Hutter}]{elsken2018neural}
\bibinfo{author}{Elsken, T.}, \bibinfo{author}{Metzen, J.H.},
  \bibinfo{author}{Hutter, F.}, \bibinfo{year}{2018}.
\newblock \bibinfo{title}{Neural architecture search: A survey}.
\newblock \bibinfo{journal}{arXiv preprint arXiv:1808.05377} .
\bibitem[{Foerster et~al.(2016)Foerster, Assael, De~Freitas and
  Whiteson}]{foerster2016learning}
\bibinfo{author}{Foerster, J.}, \bibinfo{author}{Assael, I.A.},
  \bibinfo{author}{De~Freitas, N.}, \bibinfo{author}{Whiteson, S.},
  \bibinfo{year}{2016}.
\newblock \bibinfo{title}{Learning to communicate with deep multi-agent
  reinforcement learning}, in: \bibinfo{booktitle}{Advances in neural
  information processing systems}, pp. \bibinfo{pages}{2137--2145}.
\bibitem[{Graves and Schmidhuber(2005)}]{graves2005framewise}
\bibinfo{author}{Graves, A.}, \bibinfo{author}{Schmidhuber, J.},
  \bibinfo{year}{2005}.
\newblock \bibinfo{title}{Framewise phoneme classification with bidirectional
  lstm and other neural network architectures}.
\newblock \bibinfo{journal}{Neural networks} \bibinfo{volume}{18},
  \bibinfo{pages}{602--610}.
\bibitem[{Gupta et~al.(2017)Gupta, Egorov and
  Kochenderfer}]{gupta2017cooperative}
\bibinfo{author}{Gupta, J.K.}, \bibinfo{author}{Egorov, M.},
  \bibinfo{author}{Kochenderfer, M.}, \bibinfo{year}{2017}.
\newblock \bibinfo{title}{Cooperative multi-agent control using deep
  reinforcement learning}, in: \bibinfo{booktitle}{International Conference on
  Autonomous Agents and Multiagent Systems}, \bibinfo{organization}{Springer}.
  pp. \bibinfo{pages}{66--83}.
\bibitem[{Huang et~al.(2020)Huang, Yang, Li, Qian, Huang, Shi, Dou, Chen,
  Zhang, Luo et~al.}]{huang2020searching}
\bibinfo{author}{Huang, Y.}, \bibinfo{author}{Yang, X.}, \bibinfo{author}{Li,
  R.}, \bibinfo{author}{Qian, J.}, \bibinfo{author}{Huang, X.},
  \bibinfo{author}{Shi, W.}, \bibinfo{author}{Dou, H.}, \bibinfo{author}{Chen,
  C.}, \bibinfo{author}{Zhang, Y.}, \bibinfo{author}{Luo, H.}, et~al.,
  \bibinfo{year}{2020}.
\newblock \bibinfo{title}{Searching collaborative agents for multi-plane
  localization in 3d ultrasound}, in: \bibinfo{booktitle}{International
  Conference on Medical Image Computing and Computer-Assisted Intervention},
  \bibinfo{organization}{Springer}. pp. \bibinfo{pages}{553--562}.
\bibitem[{Kaelbling et~al.(1996)Kaelbling, Littman and
  Moore}]{kaelbling1996reinforcement}
\bibinfo{author}{Kaelbling, L.P.}, \bibinfo{author}{Littman, M.L.},
  \bibinfo{author}{Moore, A.W.}, \bibinfo{year}{1996}.
\newblock \bibinfo{title}{Reinforcement learning: A survey}.
\newblock \bibinfo{journal}{Journal of artificial intelligence research}
  \bibinfo{volume}{4}, \bibinfo{pages}{237--285}.
\bibitem[{Li(2017)}]{li2017deep}
\bibinfo{author}{Li, Y.}, \bibinfo{year}{2017}.
\newblock \bibinfo{title}{Deep reinforcement learning: An overview}.
\newblock \bibinfo{journal}{arXiv preprint arXiv:1701.07274} .
\bibitem[{Li et~al.(2018)Li, Khanal et~al.}]{li2018standard}
\bibinfo{author}{Li, Y.}, \bibinfo{author}{Khanal, B.}, et~al.,
  \bibinfo{year}{2018}.
\newblock \bibinfo{title}{Standard plane detection in 3d fetal ultrasound using
  an iterative transformation network}, in: \bibinfo{booktitle}{MICCAI},
  \bibinfo{organization}{Springer}. pp. \bibinfo{pages}{392--400}.
\bibitem[{Lin et~al.(2019)Lin, Li, Ni, Liao, Wen, Du, Chen, Wang and
  Lei}]{lin2019multi}
\bibinfo{author}{Lin, Z.}, \bibinfo{author}{Li, S.}, \bibinfo{author}{Ni, D.},
  \bibinfo{author}{Liao, Y.}, \bibinfo{author}{Wen, H.}, \bibinfo{author}{Du,
  J.}, \bibinfo{author}{Chen, S.}, \bibinfo{author}{Wang, T.},
  \bibinfo{author}{Lei, B.}, \bibinfo{year}{2019}.
\newblock \bibinfo{title}{Multi-task learning for quality assessment of fetal
  head ultrasound images}.
\newblock \bibinfo{journal}{Medical image analysis} \bibinfo{volume}{58},
  \bibinfo{pages}{101548}.
\bibitem[{Liu et~al.(2018)Liu, Simonyan and Yang}]{liu2018darts}
\bibinfo{author}{Liu, H.}, \bibinfo{author}{Simonyan, K.},
  \bibinfo{author}{Yang, Y.}, \bibinfo{year}{2018}.
\newblock \bibinfo{title}{Darts: Differentiable architecture search}.
\newblock \bibinfo{journal}{arXiv preprint arXiv:1806.09055} .
\bibitem[{Lorenz et~al.(2018)Lorenz, Brosch, Ciofolo-Veit, Klinder, Lefevre,
  Cavallaro, Salim, Papageorghiou, Raynaud, Roundhill
  et~al.}]{lorenz2018automated}
\bibinfo{author}{Lorenz, C.}, \bibinfo{author}{Brosch, T.},
  \bibinfo{author}{Ciofolo-Veit, C.}, \bibinfo{author}{Klinder, T.},
  \bibinfo{author}{Lefevre, T.}, \bibinfo{author}{Cavallaro, A.},
  \bibinfo{author}{Salim, I.}, \bibinfo{author}{Papageorghiou, A.T.},
  \bibinfo{author}{Raynaud, C.}, \bibinfo{author}{Roundhill, D.}, et~al.,
  \bibinfo{year}{2018}.
\newblock \bibinfo{title}{Automated abdominal plane and circumference
  estimation in 3d us for fetal screening}, in: \bibinfo{booktitle}{Medical
  Imaging 2018: Image Processing}, \bibinfo{organization}{International Society
  for Optics and Photonics}. p. \bibinfo{pages}{105740I}.
\bibitem[{Loughna et~al.(2009)Loughna, Chitty, Evans and
  Chudleigh}]{loughna2009fetal}
\bibinfo{author}{Loughna, P.}, \bibinfo{author}{Chitty, L.},
  \bibinfo{author}{Evans, T.}, \bibinfo{author}{Chudleigh, T.},
  \bibinfo{year}{2009}.
\newblock \bibinfo{title}{Fetal size and dating: charts recommended for
  clinical obstetric practice}.
\newblock \bibinfo{journal}{Ultrasound} \bibinfo{volume}{17},
  \bibinfo{pages}{160--166}.
\bibitem[{Lu et~al.(2011)Lu, Jolly, Georgescu, Hayes, Speier, Schmidt, Bi,
  Kroeker, Comaniciu, Kellman et~al.}]{lu2011automatic}
\bibinfo{author}{Lu, X.}, \bibinfo{author}{Jolly, M.P.},
  \bibinfo{author}{Georgescu, B.}, \bibinfo{author}{Hayes, C.},
  \bibinfo{author}{Speier, P.}, \bibinfo{author}{Schmidt, M.},
  \bibinfo{author}{Bi, X.}, \bibinfo{author}{Kroeker, R.},
  \bibinfo{author}{Comaniciu, D.}, \bibinfo{author}{Kellman, P.}, et~al.,
  \bibinfo{year}{2011}.
\newblock \bibinfo{title}{Automatic view planning for cardiac mri acquisition},
  in: \bibinfo{booktitle}{International Conference on Medical Image Computing
  and Computer-Assisted Intervention}, \bibinfo{organization}{Springer}. pp.
  \bibinfo{pages}{479--486}.
\bibitem[{Maddison et~al.(2014)Maddison, Tarlow and
  Minka}]{maddison2014sampling}
\bibinfo{author}{Maddison, C.J.}, \bibinfo{author}{Tarlow, D.},
  \bibinfo{author}{Minka, T.}, \bibinfo{year}{2014}.
\newblock \bibinfo{title}{A* sampling}, in: \bibinfo{booktitle}{Advances in
  Neural Information Processing Systems}, pp. \bibinfo{pages}{3086--3094}.
\bibitem[{Mnih et~al.(2015)Mnih, Kavukcuoglu, Silver, Rusu, Veness, Bellemare,
  Graves, Riedmiller, Fidjeland, Ostrovski et~al.}]{mnih2015human}
\bibinfo{author}{Mnih, V.}, \bibinfo{author}{Kavukcuoglu, K.},
  \bibinfo{author}{Silver, D.}, \bibinfo{author}{Rusu, A.A.},
  \bibinfo{author}{Veness, J.}, \bibinfo{author}{Bellemare, M.G.},
  \bibinfo{author}{Graves, A.}, \bibinfo{author}{Riedmiller, M.},
  \bibinfo{author}{Fidjeland, A.K.}, \bibinfo{author}{Ostrovski, G.}, et~al.,
  \bibinfo{year}{2015}.
\newblock \bibinfo{title}{Human-level control through deep reinforcement
  learning}.
\newblock \bibinfo{journal}{Nature} \bibinfo{volume}{518},
  \bibinfo{pages}{529--533}.
\bibitem[{Moellers et~al.(2018)Moellers, Gr{\"u}ndahl, Hammer, Braun,
  de~Murcia, K{\"o}ster, Steinhard, Klockenbusch and
  Schmitz}]{moellers2018fetal}
\bibinfo{author}{Moellers, M.}, \bibinfo{author}{Gr{\"u}ndahl, F.},
  \bibinfo{author}{Hammer, K.}, \bibinfo{author}{Braun, J.},
  \bibinfo{author}{de~Murcia, K.O.}, \bibinfo{author}{K{\"o}ster, H.},
  \bibinfo{author}{Steinhard, J.}, \bibinfo{author}{Klockenbusch, W.},
  \bibinfo{author}{Schmitz, R.}, \bibinfo{year}{2018}.
\newblock \bibinfo{title}{Fetal brain development in diabetic pregnancies and
  normal controls}.
\newblock \bibinfo{journal}{Geburtshilfe und Frauenheilkunde}
  \bibinfo{volume}{78}, \bibinfo{pages}{P--310}.
\bibitem[{Ni et~al.(2013)Ni, Li, Yang, Qin, Li, Chin, Ouyang, Wang and
  Chen}]{ni2013selective}
\bibinfo{author}{Ni, D.}, \bibinfo{author}{Li, T.}, \bibinfo{author}{Yang, X.},
  \bibinfo{author}{Qin, J.}, \bibinfo{author}{Li, S.}, \bibinfo{author}{Chin,
  C.T.}, \bibinfo{author}{Ouyang, S.}, \bibinfo{author}{Wang, T.},
  \bibinfo{author}{Chen, S.}, \bibinfo{year}{2013}.
\newblock \bibinfo{title}{Selective search and sequential detection for
  standard plane localization in ultrasound}, in:
  \bibinfo{booktitle}{International MICCAI Workshop on Computational and
  Clinical Challenges in Abdominal Imaging}, \bibinfo{organization}{Springer}.
  pp. \bibinfo{pages}{203--211}.
\bibitem[{Ni et~al.(2014)Ni, Yang, Chen, Chin, Chen, Heng, Li, Qin and
  Wang}]{ni2014standard}
\bibinfo{author}{Ni, D.}, \bibinfo{author}{Yang, X.}, \bibinfo{author}{Chen,
  X.}, \bibinfo{author}{Chin, C.T.}, \bibinfo{author}{Chen, S.},
  \bibinfo{author}{Heng, P.A.}, \bibinfo{author}{Li, S.}, \bibinfo{author}{Qin,
  J.}, \bibinfo{author}{Wang, T.}, \bibinfo{year}{2014}.
\newblock \bibinfo{title}{Standard plane localization in ultrasound by radial
  component model and selective search}.
\newblock \bibinfo{journal}{Ultrasound in medicine \& biology}
  \bibinfo{volume}{40}, \bibinfo{pages}{2728--2742}.
\bibitem[{Real et~al.(2019a)Real, Aggarwal, Huang and Le}]{real2019aging}
\bibinfo{author}{Real, E.}, \bibinfo{author}{Aggarwal, A.},
  \bibinfo{author}{Huang, Y.}, \bibinfo{author}{Le, Q.V.},
  \bibinfo{year}{2019}a.
\newblock \bibinfo{title}{Aging evolution for image classifier architecture
  search}, in: \bibinfo{booktitle}{AAAI Conference on Artificial Intelligence}.
\bibitem[{Real et~al.(2019b)Real, Aggarwal, Huang and Le}]{real2019regularized}
\bibinfo{author}{Real, E.}, \bibinfo{author}{Aggarwal, A.},
  \bibinfo{author}{Huang, Y.}, \bibinfo{author}{Le, Q.V.},
  \bibinfo{year}{2019}b.
\newblock \bibinfo{title}{Regularized evolution for image classifier
  architecture search}, in: \bibinfo{booktitle}{Proceedings of the aaai
  conference on artificial intelligence}, pp. \bibinfo{pages}{4780--4789}.
\bibitem[{Real et~al.(2017)Real, Moore, Selle, Saxena, Suematsu, Tan, Le and
  Kurakin}]{real2017large}
\bibinfo{author}{Real, E.}, \bibinfo{author}{Moore, S.},
  \bibinfo{author}{Selle, A.}, \bibinfo{author}{Saxena, S.},
  \bibinfo{author}{Suematsu, Y.L.}, \bibinfo{author}{Tan, J.},
  \bibinfo{author}{Le, Q.V.}, \bibinfo{author}{Kurakin, A.},
  \bibinfo{year}{2017}.
\newblock \bibinfo{title}{Large-scale evolution of image classifiers}, in:
  \bibinfo{booktitle}{Proceedings of the 34th International Conference on
  Machine Learning-Volume 70}, \bibinfo{organization}{JMLR. org}. pp.
  \bibinfo{pages}{2902--2911}.
\bibitem[{Ren et~al.(2020)Ren, Xiao, Chang, Huang, Li, Chen and
  Wang}]{ren2020comprehensive}
\bibinfo{author}{Ren, P.}, \bibinfo{author}{Xiao, Y.}, \bibinfo{author}{Chang,
  X.}, \bibinfo{author}{Huang, P.Y.}, \bibinfo{author}{Li, Z.},
  \bibinfo{author}{Chen, X.}, \bibinfo{author}{Wang, X.}, \bibinfo{year}{2020}.
\newblock \bibinfo{title}{A comprehensive survey of neural architecture search:
  Challenges and solutions}.
\newblock \bibinfo{journal}{arXiv preprint arXiv:2006.02903} .
\bibitem[{Ryou et~al.(2016)Ryou, Yaqub, Cavallaro, Roseman, Papageorghiou and
  Noble}]{ryou2016automated}
\bibinfo{author}{Ryou, H.}, \bibinfo{author}{Yaqub, M.},
  \bibinfo{author}{Cavallaro, A.}, \bibinfo{author}{Roseman, F.},
  \bibinfo{author}{Papageorghiou, A.}, \bibinfo{author}{Noble, J.A.},
  \bibinfo{year}{2016}.
\newblock \bibinfo{title}{Automated 3d ultrasound biometry planes extraction
  for first trimester fetal assessment}, in: \bibinfo{booktitle}{International
  Workshop on Machine Learning in Medical Imaging},
  \bibinfo{organization}{Springer}. pp. \bibinfo{pages}{196--204}.
\bibitem[{Schaul et~al.(2015)Schaul, Quan, Antonoglou and
  Silver}]{schaul2015prioritized}
\bibinfo{author}{Schaul, T.}, \bibinfo{author}{Quan, J.},
  \bibinfo{author}{Antonoglou, I.}, \bibinfo{author}{Silver, D.},
  \bibinfo{year}{2015}.
\newblock \bibinfo{title}{Prioritized experience replay}.
\newblock \bibinfo{journal}{arXiv preprint arXiv:1511.05952} .
\bibitem[{Schlemper et~al.(2018)Schlemper, Oktay, Chen, Matthew, Knight, Kainz,
  Glocker and Rueckert}]{schlemper2018attention}
\bibinfo{author}{Schlemper, J.}, \bibinfo{author}{Oktay, O.},
  \bibinfo{author}{Chen, L.}, \bibinfo{author}{Matthew, J.},
  \bibinfo{author}{Knight, C.}, \bibinfo{author}{Kainz, B.},
  \bibinfo{author}{Glocker, B.}, \bibinfo{author}{Rueckert, D.},
  \bibinfo{year}{2018}.
\newblock \bibinfo{title}{Attention-gated networks for improving ultrasound
  scan plane detection}.
\newblock \bibinfo{journal}{arXiv preprint arXiv:1804.05338} .
\bibitem[{Schmidt-Richberg et~al.(2019)Schmidt-Richberg, Schadewaldt, Klinder,
  Lenga, Trahms, Canfield, Roundhill and Lorenz}]{schmidt2019offset}
\bibinfo{author}{Schmidt-Richberg, A.}, \bibinfo{author}{Schadewaldt, N.},
  \bibinfo{author}{Klinder, T.}, \bibinfo{author}{Lenga, M.},
  \bibinfo{author}{Trahms, R.}, \bibinfo{author}{Canfield, E.},
  \bibinfo{author}{Roundhill, D.}, \bibinfo{author}{Lorenz, C.},
  \bibinfo{year}{2019}.
\newblock \bibinfo{title}{Offset regression networks for view plane estimation
  in 3d fetal ultrasound}, in: \bibinfo{booktitle}{Medical Imaging 2019: Image
  Processing}, \bibinfo{organization}{International Society for Optics and
  Photonics}. p. \bibinfo{pages}{109493K}.
\bibitem[{Van~Hasselt et~al.(2016)Van~Hasselt, Guez and Silver}]{van2016deep}
\bibinfo{author}{Van~Hasselt, H.}, \bibinfo{author}{Guez, A.},
  \bibinfo{author}{Silver, D.}, \bibinfo{year}{2016}.
\newblock \bibinfo{title}{Deep reinforcement learning with double q-learning},
  in: \bibinfo{booktitle}{Thirtieth AAAI conference on artificial
  intelligence}.
\bibitem[{Vlontzos et~al.(2019)Vlontzos, Alansary, Kamnitsas, Rueckert and
  Kainz}]{vlontzos2019multiple}
\bibinfo{author}{Vlontzos, A.}, \bibinfo{author}{Alansary, A.},
  \bibinfo{author}{Kamnitsas, K.}, \bibinfo{author}{Rueckert, D.},
  \bibinfo{author}{Kainz, B.}, \bibinfo{year}{2019}.
\newblock \bibinfo{title}{Multiple landmark detection using multi-agent
  reinforcement learning}, in: \bibinfo{booktitle}{International Conference on
  Medical Image Computing and Computer-Assisted Intervention},
  \bibinfo{organization}{Springer}. pp. \bibinfo{pages}{262--270}.
\bibitem[{Wang et~al.(2004)Wang, Bovik, Sheikh and Simoncelli}]{wang2004image}
\bibinfo{author}{Wang, Z.}, \bibinfo{author}{Bovik, A.C.},
  \bibinfo{author}{Sheikh, H.R.}, \bibinfo{author}{Simoncelli, E.P.},
  \bibinfo{year}{2004}.
\newblock \bibinfo{title}{Image quality assessment: from error visibility to
  structural similarity}.
\newblock \bibinfo{journal}{IEEE transactions on image processing}
  \bibinfo{volume}{13}, \bibinfo{pages}{600--612}.
\bibitem[{Watkins and Dayan(1992)}]{watkins1992q}
\bibinfo{author}{Watkins, C.J.}, \bibinfo{author}{Dayan, P.},
  \bibinfo{year}{1992}.
\newblock \bibinfo{title}{Q-learning}.
\newblock \bibinfo{journal}{Machine learning} \bibinfo{volume}{8},
  \bibinfo{pages}{279--292}.
\bibitem[{Wistuba et~al.(2019)Wistuba, Rawat and Pedapati}]{wistuba2019survey}
\bibinfo{author}{Wistuba, M.}, \bibinfo{author}{Rawat, A.},
  \bibinfo{author}{Pedapati, T.}, \bibinfo{year}{2019}.
\newblock \bibinfo{title}{A survey on neural architecture search}.
\newblock \bibinfo{journal}{arXiv preprint arXiv:1905.01392} .
\bibitem[{Wong et~al.(2015)Wong, White, Ramkrishna, J{\'u}nior, Meagher and
  Costa}]{wong2015three}
\bibinfo{author}{Wong, L.}, \bibinfo{author}{White, N.},
  \bibinfo{author}{Ramkrishna, J.}, \bibinfo{author}{J{\'u}nior, E.A.},
  \bibinfo{author}{Meagher, S.}, \bibinfo{author}{Costa, F.D.S.},
  \bibinfo{year}{2015}.
\newblock \bibinfo{title}{Three-dimensional imaging of the uterus: the value of
  the coronal plane}.
\newblock \bibinfo{journal}{World journal of radiology} \bibinfo{volume}{7},
  \bibinfo{pages}{484}.
\bibitem[{Xie and Yuille(2017)}]{xie2017genetic}
\bibinfo{author}{Xie, L.}, \bibinfo{author}{Yuille, A.}, \bibinfo{year}{2017}.
\newblock \bibinfo{title}{Genetic cnn}, in: \bibinfo{booktitle}{Proceedings of
  the IEEE international conference on computer vision}, pp.
  \bibinfo{pages}{1379--1388}.
\bibitem[{Yang et~al.(2014)Yang, Ni, Qin, Li, Wang, Chen and
  Heng}]{yang2014standard}
\bibinfo{author}{Yang, X.}, \bibinfo{author}{Ni, D.}, \bibinfo{author}{Qin,
  J.}, \bibinfo{author}{Li, S.}, \bibinfo{author}{Wang, T.},
  \bibinfo{author}{Chen, S.}, \bibinfo{author}{Heng, P.A.},
  \bibinfo{year}{2014}.
\newblock \bibinfo{title}{Standard plane localization in ultrasound by radial
  component}, in: \bibinfo{booktitle}{2014 IEEE 11th International Symposium on
  Biomedical Imaging (ISBI)}, \bibinfo{organization}{IEEE}. pp.
  \bibinfo{pages}{1180--1183}.
\bibitem[{Zhang et~al.(2012)Zhang, Chen, Chin, Wang and
  Li}]{zhang2012intelligent}
\bibinfo{author}{Zhang, L.}, \bibinfo{author}{Chen, S.}, \bibinfo{author}{Chin,
  C.T.}, \bibinfo{author}{Wang, T.}, \bibinfo{author}{Li, S.},
  \bibinfo{year}{2012}.
\newblock \bibinfo{title}{Intelligent scanning: Automated standard plane
  selection and biometric measurement of early gestational sac in routine
  ultrasound examination}.
\newblock \bibinfo{journal}{Medical physics} \bibinfo{volume}{39},
  \bibinfo{pages}{5015--5027}.
\bibitem[{Zoph and Le(2016)}]{zoph2016neural}
\bibinfo{author}{Zoph, B.}, \bibinfo{author}{Le, Q.V.}, \bibinfo{year}{2016}.
\newblock \bibinfo{title}{Neural architecture search with reinforcement
  learning}.
\newblock \bibinfo{journal}{arXiv preprint arXiv:1611.01578} .
\bibitem[{Zoph et~al.(2018)Zoph, Vasudevan, Shlens and Le}]{zoph2018learning}
\bibinfo{author}{Zoph, B.}, \bibinfo{author}{Vasudevan, V.},
  \bibinfo{author}{Shlens, J.}, \bibinfo{author}{Le, Q.V.},
  \bibinfo{year}{2018}.
\newblock \bibinfo{title}{Learning transferable architectures for scalable
  image recognition}, in: \bibinfo{booktitle}{Proceedings of the IEEE
  conference on computer vision and pattern recognition}, pp.
  \bibinfo{pages}{8697--8710}.

\end{thebibliography}
	
	
	
\end{document}